\documentclass[letterpaper]{article} 
\usepackage{aaai2027}
\usepackage[hyphens]{url}  
\usepackage{graphicx} 
\usepackage{natbib}  
\usepackage{caption} 
\usepackage{booktabs}
\usepackage{subcaption} 
\usepackage{amsmath,amssymb}
\usepackage{xcolor}
\usepackage{tikz}

\usepackage{comment}
\usetikzlibrary{arrows.meta,positioning,calc,fit,backgrounds,shapes.geometric}

\IfFileExists{fontawesome5.sty}{
  \usepackage{fontawesome5}
  \newcommand{\frozenicon}{\faSnowflake}
  \newcommand{\trainicon}{\faFire}
}{
  \newcommand{\frozenicon}{$\ast$}
  \newcommand{\trainicon}{$\blacktriangle$}
}

\title{A Glance Is All You Need:\\Single-Pass Fine-Grained Image Captioning with SimLoss}

\author{
Suryaansh~Jain\textsuperscript{\rm 1, \rm 3},
Rahasya~Barkur\textsuperscript{\rm 1},
Vishal~G\textsuperscript{\rm 1},
Ryan~Rossi\footnote{This work was done when the author was at Adobe Research.},
Franck~Dernoncourt\textsuperscript{\rm 2},
Jack~Wang\textsuperscript{\rm 2},\\
Koustava~Goswami\textsuperscript{\rm 2},
Nedim~Lipka\textsuperscript{\rm 2},
Puneet~Mathur\textsuperscript{\rm 2},
Samyadeep~Basu\textsuperscript{\rm 2},
Seunghyun~Yoon\textsuperscript{\rm 2}
}

\affiliations{
\textsuperscript{\rm 1}University of Massachusetts Amherst\\
\textsuperscript{\rm 2}Adobe Research\\
\textsuperscript{\rm 3}Applied Materials
}

\begin{document}
\maketitle

\begin{abstract}
An image may be worth a thousand words, but most captioning models describe it in only a few. Modern vision-language models produce fluent high-level captions, yet routinely miss the attributes, counts, textures, materials, and spatial relations that make an image visually specific.
Recent multi-stage systems recover some of these details through generation, decomposition, verification, and rewriting, but they do so at the expense of substantially higher inference latency.

We propose \emph{SimLoss}, a reference-free embedding-space objective for single-pass fine-grained image captioning. SimLoss trains a vision-language model to align its projected hidden-state representation with a frozen image embedding through an InfoNCE contrastive loss, supplying a dense visual supervision signal before any text is decoded, and requiring neither human-written fine-grained captions nor pseudo-captions from a multi-stage pipeline. We instantiate it as \emph{SimLoss FFT}, which backpropagates through a locally available embedding model, and \emph{SimLoss GRPO}, which treats that model as a black-box reward.

Compared with single-pass, multi-stage verification, reward-optimized, and perception-aware baselines, the fully differentiable fine-tuning variant, \emph{SimLoss FFT}, achieves the highest precision while nearly matching the F1 score of the multi-stage method, all while retaining single-pass inference and running roughly $20\times$ faster than the multi-stage pipeline. 
The reward-based variant \emph{SimLoss GRPO} attains the strongest recall. Together, these results show that embedding-space supervision can recover the quality of multi-stage verification at the latency of a single-pass captioner. Code can be found at  https://github.com/srynsh/SimLoss-Image-Captioning

\end{abstract}

\section{Introduction}

\begin{figure}[t!]
\centering
\includegraphics[width=0.46\textwidth]{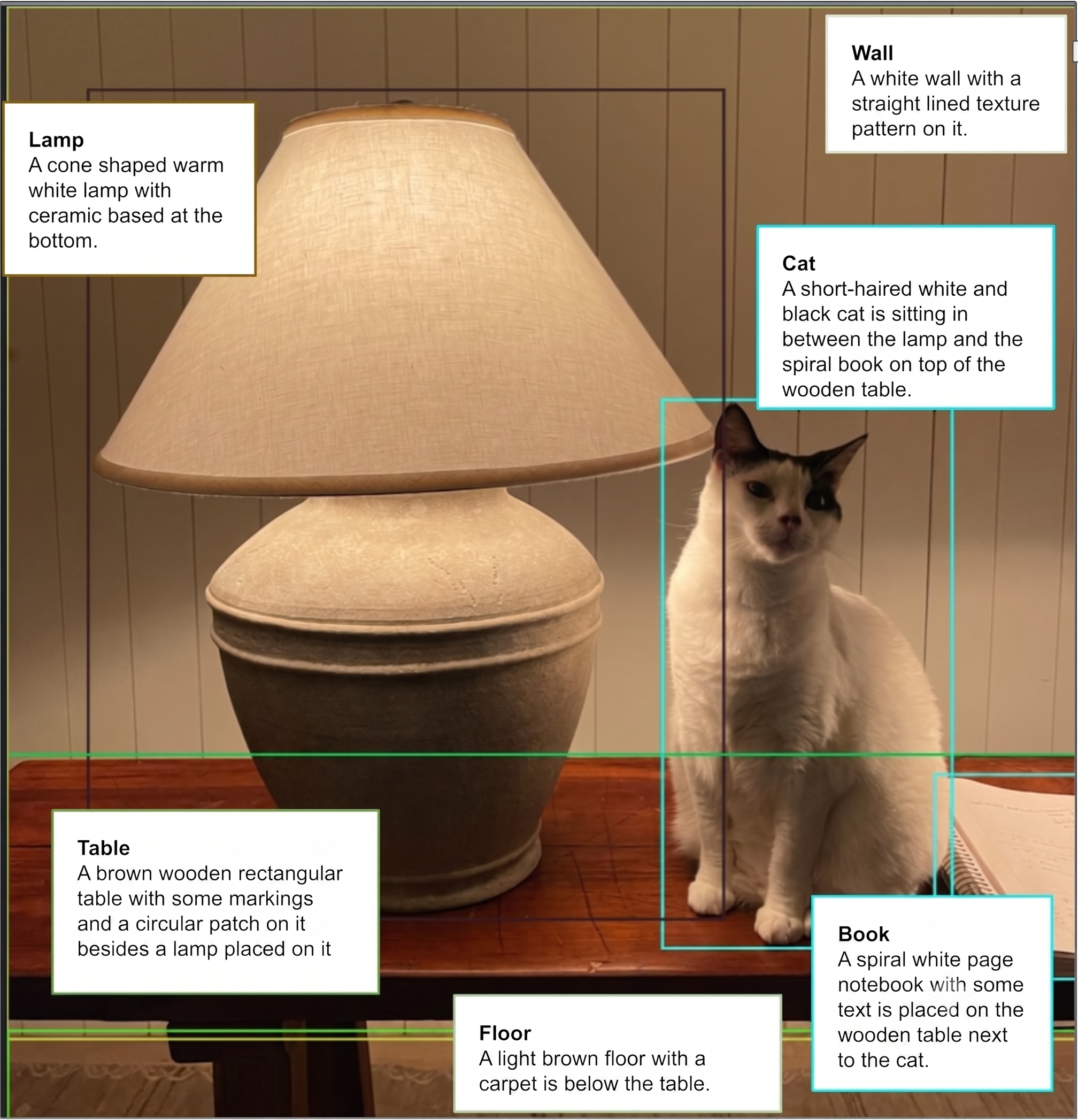}
\caption{
Example of a human-written fine-grained caption. Fine-grained captioning requires visually specific details. A generic caption may identify the main objects while omitting attributes, materials, counts, textures, and spatial relationships that distinguish the image.
}
\label{fig:iiw_example}
\vspace{-6pt}
\end{figure}

Humans can glance at an image once and retain fine visual details: the material of a lamp base, the pattern on a fabric, the number of repeated objects, and the spatial relations among small items. Modern vision-language models (VLMs), despite their fluency, often miss this level of specificity. They may produce captions that are broadly correct but visually incomplete, such as describing Figure~\ref{fig:iiw_example} as ``a lamp on a table'' while omitting the urn-shaped ceramic base, the carved rings along the stem, and the spiral-bound notebook beside it.

These omissions limit the usefulness of captions in settings where grounded detail matters, including assistive technology, embodied robotics, and clinical image interpretation~\citep{ahsan2021multimodalvisuallyimpaired,brohan2023rt2,jing2018automaticgenerationmedical}. The challenge is not fluency, but faithful coverage. Standard captioning datasets such as MS~COCO emphasize concise descriptions of salient objects~\citep{lin2015microsoftcococommonobjects}, while asking models for more detail can increase hallucination~\citep{li2023pope,leng2024vcd,zhou2024lure}. Recent hyper-detailed captioning benchmarks therefore focus on the precision--coverage tradeoff: useful captions should recover fine visual evidence without adding unsupported claims~\citep{garg2024imageinwordsunlockinghyperdetailedimage,onoe2024docci,lee2025robusthyperdetailedimagecaptioning,ye2025paintingwordselevatingdetailed}.

A common solution is to add inference-time computation. CapMAS~\citep{lee2025robusthyperdetailedimagecaptioning} generates multiple captions, decomposes them into atomic claims, verifies each claim against the image, and rewrites the caption using verified content. Patch-based and region-aware methods similarly add local perception or aggregation stages to recover details missed by a single global pass~\citep{peng2025patchmatterstrainingfreefinegraine}. These approaches improve factuality or coverage, but they turn captioning into a multi-pass pipeline, which is costly for interactive and large-scale use. We ask whether this benefit can instead be moved into training without collecting fine-grained captions or generating them with a teacher pipeline. Our key idea is to view a fine-grained captioner as preserving discriminative visual information. A generic caption remains compatible with many similar images, while a detailed caption should make the source image easier to identify. This suggests an embedding-space training signal: align the VLM representation produced from an image with a frozen image embedding that already captures visual similarity.


We introduce \emph{SimLoss}, a reference-free objective for single-pass fine-grained captioning. During training, a frozen multimodal embedding model encodes the image while a trainable VLM processes the same image and prompt; we mean-pool the VLM hidden states, project them into the embedding space, and align them contrastively with the frozen image embedding. Because the adapter weights are shared between this pooling path and the generation path, the objective reshapes the representations the decoder reads from without ever prescribing wording. At inference the embedding model and projector are removed, so captioning uses only the adapted VLM in a single pass. SimLoss therefore needs no caption targets at all, and when the embedding model is locally available it supplies a fully differentiable grounding signal before any discrete text is sampled.


Our contributions are:
\begin{itemize}
\itemsep0em
\item SimLoss, a reference-free contrastive objective that supervises a captioner in embedding space before decoding, motivated by a noisy-channel view of captioning (Section~\ref{sec:formulation}).
\item Two instantiations covering the differentiable and black-box settings, \emph{SimLoss FFT} and \emph{SimLoss GRPO} (Sections~\ref{sec:simloss_fft},~\ref{sec:simloss_grpo}), which align different objects and consequently trade precision against coverage differently.
\item An evaluation on IIW-400 against single-pass, multi-stage verification, reward-optimized, and perception-aware baselines. SimLoss FFT is the most precise method we evaluate and is indistinguishable from CapMAS in F1 at roughly $20\times$ lower latency. We also find recall essentially flat across all methods, which localizes the entire F1 spread to precision.
\end{itemize}
\section{Related Work}

Image captioning has evolved from encoder--decoder models with visual attention~\citep{vinyals2015showtell,xu2015showattendtell} to instruction-tuned vision-language models~\citep{liu2023visualinstructiontuning,li2023blip2}, but fine-grained captioning remains difficult. Early controllable captioning methods encouraged descriptions of attributes, relations, and scene structure beyond salient objects~\citep{zha2019contextaware,chen2020sayasyouwish}. Recent long-form captioning benchmarks such as ImageInWords and DOCCI show that even strong VLMs remain incomplete or inconsistent when asked for dense visual descriptions~\citep{garg2024imageinwordsunlockinghyperdetailedimage,onoe2024docci}. Several methods address this by adding inference-time computation. CapMAS decomposes captions into atomic claims, verifies them, and rewrites the caption using supported content~\citep{lee2025robusthyperdetailedimagecaptioning}; VisualFactChecker similarly uses external detection and visual question-answering tools for post-hoc correction~\citep{ge2024visualfactchecker}; and Patch Matters recovers local detail through patch-level aggregation~\citep{peng2025patchmatterstrainingfreefinegraine}. These methods target the same precision--coverage problem as ours, but typically improve caption quality by increasing inference cost.

A related line uses image-text similarity, retrieval, or rewards as supervision. CLIPScore introduced reference-free caption evaluation using image-text similarity~\citep{hessel2021clipscore}, and fine-grained captioning with CLIP reward used this signal to encourage distinctive captions without relying only on reference captions~\citep{cho2022finegrainedclipreward}. Self-retrieval objectives are closely related, although naive optimization can reduce faithfulness and increase hallucination, motivating methods such as Visual Caption Boosting and BagCurri~\citep{gaur2024nodetailleftbehind}. Hallucination has also been studied through object-level evaluation and mitigation methods such as POPE, contrastive decoding, beam-search penalties, and post-hoc correction~\citep{li2023pope,leng2024vcd,huang2024opera,zhou2024lure,yin2023woodpecker}. Recent reward-optimized and perception-aware captioners, including FeedQuill and PAPO, further optimize detailed caption quality with learned or structured feedback~\citep{ye2025paintingwordselevatingdetailed,wang2025perceptionawarepolicyoptimizationmultimodal}. SimLoss shares the similarity-as-supervision intuition of this work, but applies the signal before decoding by aligning the captioner's continuous representation with a frozen image embedding, rather than scoring sampled captions after generation.

Our objective is contrastive rather than distillative: no teacher caption distribution is matched~\citep{hinton2015distillingknowledgeneuralnetwork}. It reuses the InfoNCE loss~\citep{oord2018representation} underpinning contrastive vision--language pretraining~\citep{radford2021learning}, but applies it between a frozen embedding model and the pooled hidden state of a captioner being adapted, so the signal shapes generation rather than a retrieval encoder.


\section{Problem and Motivation}
\label{sec:problem}

\subsection{Detailed Captioning Requires Coverage and Grounding}

Figure~\ref{fig:iiw_example} illustrates the core difficulty. A caption such as ``a cat sitting on a table near a lamp'' is not wrong, but it is incomplete. It identifies the dominant objects while omitting much of the visual evidence that makes the image specific. The cat's markings, the ceramic structure of the lamp base, the spiral-bound notebook, the wood grain of the table, and the arrangement of the objects are all visible, yet they are easy for a fluent vision-language model to omit.

We use \emph{fine-grained captioning} to refer to captions that preserve discriminative visual information. A useful caption should include attributes, counts, textures, materials, object parts, and spatial relations when they are visible in the image. It should also avoid adding plausible but unsupported details. The task is therefore not simply to generate longer captions. It is to improve visual coverage while maintaining factual grounding.

\subsection{Caption Sources Differ in Detail and Reliability}
\label{sec:datasets}




Standard caption supervision is not designed for this setting. MS~COCO provides over 120{,}000 images with five human captions per image~\citep{lin2015microsoftcococommonobjects}, effective for learning generic scene description but intentionally concise: in our sampled batch, COCO captions average 10.0 words, rarely exhausting attributes, materials, textures, or fine spatial layout. ImageInWords studies the more demanding setting of hyper-detailed description and provides IIW-400 as an evaluation set~\citep{garg2024imageinwords}; its human descriptions average 171.2 words, roughly $17\times$ longer. We use IIW-400 for evaluation only, never as a training source.

Multi-stage pipelines can also produce detailed captions, but not clean supervision. In our processed IIW-400 outputs, the verified final captions from CapMAS (Section~\ref{sec:baselines}) average 186.0 words and 29.3 atomic propositions, of which 22.3 are judged true and 7.0 false, a mean factuality ratio of 0.766. These statistics explain both the appeal of pipeline captions and the risk of imitating them: they carry far richer detail than ordinary supervision, but roughly a quarter of their propositions do not survive verification, so caption-level distillation would transfer their unverifiable claims and stylistic artifacts along with their detail. This is the supervision problem SimLoss sidesteps.

\subsection{Evaluation Measures}
\label{sec:evaluation_measures}

We follow the CapMAS dual protocol~\citep{lee2025robusthyperdetailedimagecaptioning}. \emph{Precision} (factuality) decomposes a caption into atomic propositions and reports the fraction a multimodal judge finds supported by the image and IIW reference description. \emph{Recall} (coverage) is the fraction of human-verified, image-derived multiple-choice questions answered correctly from the caption alone, with the image withheld. F1 is their harmonic mean. We additionally report CLAIR (\emph{Criterion using LAnguage models for Image caption Rating}), a reference-based LLM metric that rates how likely the candidate and human-reference captions describe the same image~\citep{chan-etal-2023-clair-evaluating}; we divide its 0--100 output by 100. Caption length and latency measure verbosity and efficiency. The IIW references are used only for evaluation, never as SimLoss adaptation targets. Formal definitions and full prompts appear in Appendix~\ref{sec:prompts_eval}.

\subsection{Captions as Noisy Channels}
\label{sec:formulation}

We view fine-grained captioning through an information-theoretic lens. Let $v$ be an input image, let $x$ be a captioning prompt, and let $\hat{y} \sim \pi_\theta(\cdot \mid x, v)$ be the generated caption. A caption is a lossy textual channel from the image to language. If the caption preserves only coarse scene information, many visually distinct images remain compatible with it. If it preserves fine-grained evidence, such as attributes, counts, materials, textures, and spatial relations, the uncertainty about the source image should be lower.

This suggests an ideal objective in terms of mutual information. A better caption should retain more information about the image, which means maximizing
\begin{equation}
I(v; \hat{y})
=
H(v) - H(v \mid \hat{y}).
\end{equation}
Since $H(v)$ is fixed for the data distribution, the goal is equivalently to reduce the residual uncertainty $H(v \mid \hat{y})$. Intuitively, a caption is better when it makes the source image easier to distinguish from other plausible alternatives.

Standard supervised captioning does not optimize this objective directly. Instead, it uses token-level cross-entropy against a reference caption $y^{*} = (y^{*}_{1}, \dots, y^{*}_{T})$:
\begin{equation}
\mathcal{L}_{\mathrm{CE}}
=
- \sum_{t=1}^{T}
\log \pi_{\theta}(y^{*}_{t} \mid y^{*}_{<t}, x, v).
\end{equation}
This objective is effective when the reference caption is an adequate target. For fine-grained captioning, however, the reference caption is itself only one lossy projection of the image. A short or incomplete caption can still achieve low cross-entropy even when it omits visually present details. Cross-entropy therefore encourages the model to imitate a particular caption realization, including its omissions, rather than to preserve all image-specific evidence.

Conventional training uses weak caption supervision from MS~COCO, while evaluation targets the much richer descriptive regime represented by IIW. SimLoss addresses this mismatch using COCO images but not their captions, IIW descriptions, or CapMAS captions as adaptation targets.

Directly maximizing $I(v; y)$ is difficult because captions are discrete sequences and mutual information over image-text pairs is not directly tractable during caption generation. SimLoss therefore applies the same principle before decoding, at the level of continuous representations. Let $E_I$ be a frozen image encoder that maps images into $\mathbb{R}^{d}$. Let $h_\theta(v,x) \in \mathbb{R}^{d_h}$ denote the VLM hidden-state representation, and let $W \in \mathbb{R}^{d \times d_h}$ be a learned projector. We define
\begin{equation}
z_{\mathrm{img}} = E_I(v),
\end{equation}
\begin{equation}
z_{\mathrm{vlm}} = W h_\theta(v,x).
\end{equation}
SimLoss optimizes a representation-level proxy that encourages $z_{\mathrm{vlm}}$ to retain enough information to identify $z_{\mathrm{img}}$ among alternatives. A generic representation should match many images. A representation that preserves fine-grained evidence should retrieve the source image more reliably. We turn this intuition into a contrastive training objective in Section~\ref{sec:simloss}.

\subsection{Embedding-Space Similarity Across Caption Sources}
\label{sec:embedding_similarity}

The noisy-channel view suggests a simple diagnostic. If a caption preserves image-specific information, it should align more strongly with the source image and with detailed human descriptions than a generic baseline caption does. We study this using a frozen Qwen3-VL-Embedding model.

For each image $v_i$, we compute a shared-space embedding for the available sources in each dataset. For IIW, the source set is
\begin{equation}
\mathcal{S}_{\mathrm{IIW}}
=
\{\mathrm{Image}, \mathrm{IIW}, \mathrm{CapMAS}, \mathrm{Base}\},
\end{equation}
where \emph{Image} is the ground-truth image embedding, \emph{IIW} is the human hyper-detailed description, \emph{CapMAS} is the verified CapMAS final caption, and \emph{Base} is the baseline Qwen2.5-VL-7B generation. For MS~COCO, the source set is
\begin{equation}
\mathcal{S}_{\mathrm{COCO}}
=
\{\mathrm{Image}, \mathrm{COCO}, \mathrm{Base}\},
\end{equation}
where \emph{COCO} is the human caption and \emph{Base} is again the baseline Qwen2.5-VL-7B generation.

For source types $a$ and $b$ in the relevant source set, let $e_i^{a}$ and $e_i^{b}$ denote the corresponding embeddings for example $i$. We define the mean matched-pair similarity
\begin{equation}
M_{ab}
=
\frac{1}{N}\sum_{i=1}^{N}\cos(e_i^{a}, e_i^{b}).
\end{equation}
This produces a cross-source similarity matrix for each dataset.

COCO characterizes the conventional weak-caption regime, while IIW describes the richer fine-grained regime that the model must generalize to at evaluation time. SimLoss uses the COCO images in adaptation but discards their captions.

The strongest contrast appears in the COCO setting. The ground-truth image aligns only weakly with the COCO human caption, with mean similarity 0.4794, while its similarity to the baseline Qwen2.5-VL-7B generation is 0.6982. This gap reflects how coarse ordinary caption supervision is relative to the richer semantic structure captured by the shared embedding space. The COCO human caption is also only moderately aligned with the baseline generation, with similarity 0.5055. Ground-truth images align with IIW human descriptions at 0.6616, with CapMAS final captions at 0.7000, and with baseline generations at 0.6982. CapMAS and the baseline are therefore both close to the source image under this diagnostic. The more informative comparison is between automated captions and detailed human text. CapMAS final captions align with IIW human descriptions at 0.6928, while baseline generations align with IIW human descriptions at 0.6803. This suggests that CapMAS moves slightly closer to the detailed-description regime, even though both automated caption sources are already much closer to that regime than ordinary COCO supervision is.

These results sharpen the motivation for SimLoss. The problem is not that the base captioner is disconnected from the image; under this diagnostic its generations already sit close to the source image. The problem is the gap between what supervision provides and what evaluation demands: short COCO captions during training versus dense, grounded description on IIW. SimLoss closes this gap with the image embedding itself, rather than with fine-grained human captions or pipeline-generated captions as targets.

\begin{table}[t]
\centering
\small
\renewcommand{\arraystretch}{1.12}
\begin{tabular}{lccc}
\hline
IIW-400 & IIW human & CapMAS & Qwen2.5-VL \\
\hline
Image & 0.6616 & 0.7000 & 0.6982 \\
IIW human & --- & 0.6928 & 0.6803 \\
CapMAS &  & --- & 0.7307 \\
\hline
\end{tabular}
\caption{Cross-source embedding similarity on IIW-400. Entries are mean matched-pair cosine similarities in a frozen Qwen3-VL-Embedding space.}
\label{tab:iiw_cross_source_similarity}
\end{table}

\begin{table}[t]
\centering
\small
\renewcommand{\arraystretch}{1.12}
\begin{tabular}{lcc}
\hline
MS~COCO & COCO human & Qwen2.5-VL\\
\hline
Image & 0.4794 & 0.6982 \\
COCO human & --- & 0.5055 \\
\hline
\end{tabular}
\caption{Cross-source embedding similarity on MS~COCO. Entries are mean matched-pair cosine similarities in a frozen Qwen3-VL-Embedding space.}
\label{tab:coco_cross_source_similarity}
\end{table}

\begin{figure*}[t]
\centering
\resizebox{\textwidth}{!}{%
\begin{tikzpicture}[
    x=1cm,y=1cm,
    >=Latex,
    font=\sffamily,
    fwd/.style={-{Latex[length=2.1mm,width=1.7mm]}, line width=0.64pt, draw=black!90},
    gradline/.style={dashed, line width=0.74pt, draw=simpurple!92},
    back/.style={-{Latex[length=2.1mm,width=1.7mm]}, dashed, line width=0.74pt, draw=simpurple!92},
    box/.style={rounded corners=4pt, line width=0.62pt, align=center,
                inner xsep=4.2pt, inner ysep=3.5pt, font=\scriptsize\sffamily},
    paneltitle/.style={font=\bfseries\sffamily\Large},
    panelsubtitle/.style={font=\scriptsize\sffamily},
    edgeword/.style={font=\tiny\sffamily, fill=simbluefill, inner sep=1.2pt},
    legendtxt/.style={font=\scriptsize\sffamily},
]

\definecolor{simblue}{RGB}{24,94,205}
\definecolor{simbluefill}{RGB}{246,249,255}
\definecolor{simgreen}{RGB}{56,137,64}
\definecolor{simgreenfill}{RGB}{248,253,248}
\definecolor{simpurple}{RGB}{110,58,214}
\definecolor{simpurplefill}{RGB}{251,248,255}
\definecolor{simorange}{RGB}{235,146,26}
\definecolor{simorangefill}{RGB}{255,250,240}
\definecolor{simred}{RGB}{204,54,48}
\definecolor{simredfill}{RGB}{255,248,247}
\definecolor{simgray}{RGB}{112,119,128}
\definecolor{simgrayfill}{RGB}{250,250,250}
\definecolor{titleblack}{RGB}{8,18,35}

\tikzset{
  frozenbox/.style={box, draw=simblue, fill=simbluefill},
  parambox/.style={box, draw=simpurple, fill=simpurplefill},
  mixedbox/.style={box, draw=simpurple, fill=white},
  opbox/.style={box, draw=simgray!65, fill=simgrayfill},
  inputbox/.style={box, draw=simgray!70, fill=white},
  inferbox/.style={box, draw=simgreen!78!black, fill=simgreenfill},
  lossbox/.style={box, draw=simorange, fill=simorangefill},
}

\coordinate (trainSW) at (0.10,0.75);
\coordinate (trainNE) at (17.05,7.10);
\coordinate (inferSW) at (17.35,0.75);
\coordinate (inferNE) at (20.50,7.10);

\begin{scope}[on background layer]
  \node[draw=simgray!34, fill=simbluefill, rounded corners=9pt, line width=0.62pt,
        fit=(trainSW)(trainNE), inner sep=0pt] (trainpanel) {};
  \node[draw=simgreen!40, fill=simgreenfill, rounded corners=9pt, line width=0.62pt,
        fit=(inferSW)(inferNE), inner sep=0pt] (inferpanel) {};
\end{scope}

\node[font=\bfseries\sffamily\Large, text=titleblack] at (10.30,7.72)
{SimLoss: Fully Differentiable Fine-Tuning};

\node[paneltitle, text=simblue] at (8.58,6.78) {TRAINING};
\node[draw=simorange!58, fill=simorangefill, rounded corners=3pt,
      font=\scriptsize\bfseries\sffamily, text=titleblack,
      inner xsep=7pt, inner ysep=2.8pt] at (8.58,6.29)
{No caption targets: supervision comes from image embeddings and batch identities};

\node[paneltitle, text=simgreen!82!black] at (18.93,6.78) {INFERENCE};
\node[panelsubtitle, text=simgreen!75!black] at (18.93,6.35)
{single-pass generation};

\node[inputbox, minimum width=1.48cm, minimum height=1.04cm, text width=1.18cm]
  (images) at (1.02,4.85)
  {Image batch\\[-1pt] $\{v_i\}_{i=1}^{N}$};

\node[frozenbox, minimum width=2.30cm, minimum height=1.04cm, text width=2.00cm]
  (imgenc) at (3.42,4.85)
  {{\color{simblue}\normalsize\frozenicon}\;\; Frozen $E_I$\\[-1pt]
   Qwen3-VL-Embed};

\node[opbox, minimum width=1.86cm, minimum height=1.04cm, text width=1.56cm]
  (imgemb) at (6.28,4.85)
  {Image target\\[-1pt] $z_i^I=E_I(v_i)$};

\node[inputbox, minimum width=1.48cm, minimum height=1.12cm, text width=1.18cm]
  (prompt) at (1.02,2.75)
  {Fixed prompt\\[-1pt] $x$};

\node[mixedbox, minimum width=2.18cm, minimum height=1.12cm, text width=1.88cm]
  (vlm) at (3.42,2.75)
  {\textbf{Qwen2.5-VL}\\[-2pt]
   {\color{simblue}\tiny \frozenicon\; frozen base}\\[-2pt]
   {\color{simorange}\tiny \trainicon}\;{\color{simpurple}\tiny trainable LoRA $\Delta\theta$}};

\node[opbox, minimum width=1.44cm, minimum height=1.12cm, text width=1.14cm]
  (states) at (5.58,2.75)
  {Token states\\[-1pt] {\tiny $\{H_{i,t}\}$}};

\node[opbox, minimum width=2.18cm, minimum height=1.12cm, text width=1.88cm]
  (pool) at (7.74,2.75)
  {\textbf{Mean pool}\\[-1pt]
   {\tiny $h_\theta(v_i,x)=T_i^{-1}\sum_t H_{i,t}$}};

\node[parambox, minimum width=1.36cm, minimum height=1.12cm, text width=1.06cm]
  (projector) at (9.86,2.75)
  {{\color{simorange}\small\trainicon}\\[-2pt]
   Projector $g_\phi$};

\node[opbox, minimum width=1.34cm, minimum height=1.12cm, text width=1.04cm]
  (vlmemb) at (11.56,2.75)
  {VLM output\\[-1pt] $z_j^V$};

\node[opbox, minimum width=2.50cm, minimum height=1.24cm, text width=2.20cm]
  (contrast) at (13.85,3.80)
  {\textbf{Identity contrast}\\[-2pt]
   {\tiny $s_{ij}=\cos(z_i^I,z_j^V)$}\\[1pt]
   {\color{simgreen!82!black}\tiny $i=j$: $s_{ii}$ (positive)}\\[-1pt]
   {\color{simred}\tiny $j\ne i$: $s_{ij}$ (negative)}};

\node[lossbox, minimum width=1.34cm, minimum height=1.02cm, text width=1.04cm]
  (loss) at (16.15,3.80)
  {\textbf{InfoNCE}\\[-1pt] {\tiny $\mathcal{L}_{\mathrm{SimLoss}}$}};

\draw[fwd] (images.east) -- node[edgeword, above] {$v_i$} (imgenc.west);
\draw[fwd] (imgenc.east) -- (imgemb.west);

\coordinate (imageToVLMTop) at ($(images.south)+(0,-0.34)$);
\coordinate (imageToVLMRight) at (vlm.north |- imageToVLMTop);
\draw[fwd] (images.south) -- (imageToVLMTop)
  -- node[edgeword, above] {$v_i$} (imageToVLMRight) -- (vlm.north);
\draw[fwd] (prompt.east) -- node[edgeword, above] {$x$} (vlm.west);

\draw[fwd] (vlm.east) -- (states.west);
\draw[fwd] (states.east) -- (pool.west);
\draw[fwd] (pool.east) -- (projector.west);
\draw[fwd] (projector.east) -- (vlmemb.west);

\coordinate (teacherJoin) at (contrast.north |- imgemb.east);
\coordinate (studentJoin) at (contrast.south |- vlmemb.east);
\draw[fwd] (imgemb.east) -- (teacherJoin) -- (contrast.north);
\draw[fwd] (vlmemb.east) -- (studentJoin) -- (contrast.south);
\draw[fwd] (contrast.east) -- (loss.west);

\coordinate (gradBusR) at (loss.south |- 0,1.10);
\coordinate (gradBusL) at (vlm.south |- 0,1.10);
\coordinate (gradProjector) at (projector.south |- 0,1.10);
\draw[gradline] (loss.south) -- (gradBusR) -- (gradBusL);
\draw[back] (gradBusL) -- (vlm.south);
\draw[back] (gradProjector) -- (projector.south);
\node[font=\scriptsize\itshape\sffamily, text=simpurple!92, fill=simbluefill,
      inner sep=1.4pt] at (6.65,1.40)
{gradients update only LoRA $\Delta\theta$ and $g_\phi$};

\node[inferbox, minimum width=2.16cm, minimum height=0.82cm, text width=1.86cm]
  (infinput) at (18.93,5.55)
  {Image $v$ + prompt $x$};

\node[inferbox, minimum width=2.16cm, minimum height=0.82cm, text width=1.86cm]
  (infmodel) at (18.93,4.30)
  {Qwen2.5-VL +\\[-1pt] learned LoRA};

\node[inferbox, minimum width=2.16cm, minimum height=0.82cm, text width=1.86cm]
  (caption) at (18.93,3.05)
  {Single-pass caption\\[-1pt] $\hat y$};

\node[box, draw=simgray!48, fill=white, dashed, text=simgray, font=\tiny\sffamily,
      minimum width=2.16cm, minimum height=0.72cm, text width=1.86cm,
      inner xsep=3pt, inner ysep=2.4pt]
  (removed) at (18.93,1.55)
  {Removed after training:\\[-1pt] $E_I$, $g_\phi$, $\mathcal{L}_{\mathrm{SimLoss}}$};

\draw[fwd] (infinput.south) -- (infmodel.north);
\draw[fwd] (infmodel.south) -- (caption.north);

\node[text=simblue, font=\large] at (3.45,0.18) {\frozenicon};
\node[legendtxt, anchor=west] at (3.76,0.18) {Frozen};

\node[text=simorange, font=\large] at (6.72,0.18) {\trainicon};
\node[legendtxt, anchor=west] at (7.03,0.18) {Trainable};

\draw[fwd] (10.00,0.18) -- (10.66,0.18);
\node[legendtxt, anchor=west] at (10.84,0.18) {Forward computation};

\draw[back] (14.50,0.18) -- (15.16,0.18);
\node[legendtxt, anchor=west] at (15.34,0.18) {Gradient path};

\end{tikzpicture}%
}
\caption{\textbf{SimLoss fully differentiable fine-tuning (FFT).}
The frozen encoder produces image targets $z_i^I$, while Qwen2.5-VL's token
states are mean-pooled and projected to $z_j^V$.  Cosine scores for all batch
pairs are separated explicitly into matched identities ($s_{ii}$, green) and
other identities ($s_{ij}$ for $j\ne i$, red) before entering InfoNCE.  The loss
updates only LoRA and the projector, without caption targets.  At inference,
the encoder, projector, and contrastive loss are removed.}
\label{fig:simloss}
\end{figure*}
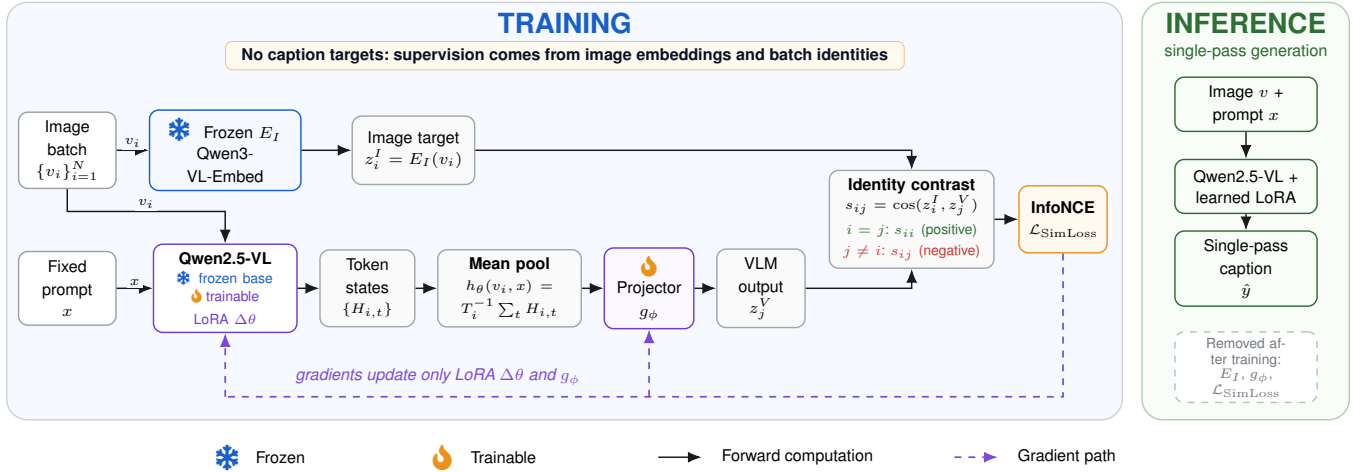
\section{Method}
\label{sec:method}

\subsection{SimLoss: Embedding-Space Distillation}
\label{sec:simloss}

Figure~\ref{fig:simloss} summarizes SimLoss training and inference. Standard caption-level supervision requires a target caption, and reward-based methods require sampling discrete text; both routes are limited by either pseudo-label quality or high-variance policy gradients. SimLoss instead operates on continuous VLM representations. An ideal detailed caption preserves image information, motivating the mutual-information view $I(\text{img};\text{caption}) = H(\text{img}) - H(\text{img} \mid \text{caption})$. Because this quantity is intractable over discrete captions, we use pre-decoding representation alignment as a proxy: a frozen embedding space supplies an image target, and the VLM's pooled hidden state is trained to distinguish its source image from in-batch alternatives. Let $v_i$ be an image and $x$ be the captioning prompt. Let $E_I$ denote a frozen image encoder, instantiated as Qwen3-VL-Embed, which maps the image to an embedding
\begin{equation}
z_i^I = E_I(v_i) \in \mathbb{R}^{d}.
\end{equation}
Let $h_\theta(v_i,x) \in \mathbb{R}^{d_h}$ denote the pooled hidden-state representation produced by Qwen2.5-VL-7B. We attach a learned projector $g_\phi: \mathbb{R}^{d_h} \rightarrow \mathbb{R}^{d}$ and define
\begin{equation}
z_i^V = g_\phi(h_\theta(v_i,x)).
\end{equation}
In our implementation, $g_\phi$ is a two-layer multilayer perceptron (MLP) that maps the VLM hidden dimension to the Qwen3-VL-Embed image-embedding dimension. Given a batch of $N$ image-prompt pairs, we compute cosine similarities
\begin{equation}
s_{ij} = \cos(z_i^I, z_j^V),
\end{equation}
where $z_i^I$ is the frozen embedding of image $i$ and $z_j^V$ is the projected VLM representation for image $j$. We optimize the InfoNCE loss~\citep{oord2019representationlearningcontrastivepredictive}
\begin{equation}
\mathcal{L}_{\mathrm{SimLoss}}
= -\frac{1}{N}\sum_{i=1}^{N}
\log
\frac{\exp(s_{ii}/\tau)}
{\sum_{j=1}^{N}\exp(s_{ij}/\tau)},
\label{eq:simloss}
\end{equation}
where $\tau$ is a temperature hyperparameter. This objective treats the correct image-representation pair as the positive pair and all other images in the batch as negatives. A generic VLM representation that discards visual details should be harder to match uniquely to the source image. A representation that preserves fine-grained attributes, counts, textures, materials, and spatial relations should align more strongly with the correct image embedding. Thus, SimLoss encourages the captioning model to retain discriminative visual information before text generation occurs. Unlike caption-level distillation, SimLoss does not require human-written fine-grained captions or CapMAS pseudo-captions as training targets. Unlike reinforcement-learning (RL) caption optimization, it does not require sampling captions during training when the embedding model is differentiable. The training path through the low-rank adaptation (LoRA) modules, pooled hidden state, and projector is differentiable, so gradients update the trainable VLM adapters directly.

\subsection{SimLoss Fully Differentiable Fine-Tuning (FFT)}
\label{sec:simloss_fft}

We call the differentiable projector-alignment variant SimLoss FFT. Here, \emph{fully differentiable} describes the training signal, not full-parameter tuning: gradients pass through the projector into the LoRA adapters, while the base weights and embedding encoder remain frozen. For each image, we compute $z_i^I$ with the locally accessible frozen image encoder and $z_i^V$ from the VLM hidden states, then optimize Equation~\ref{eq:simloss} over in-batch positives and negatives. During inference, the projector and frozen encoder are removed, leaving the original single-pass structure.

\subsection{SimLoss GRPO with Black-Box Embeddings}
\label{sec:simloss_grpo}

The projector-based SimLoss objective requires access to the embedding model's internal computation graph. When the embedding model is closed-source or accessible only through an API, direct backpropagation is not possible. To handle this setting, we define a reward-based variant, SimLoss GRPO. Given an image $v$ and a sampled caption $\hat{y} \sim \pi_\theta(\cdot \mid x,v)$, we encode the image and text using a black-box multimodal embedding model:
\begin{equation}
z^I = E_I(v), \qquad z^T = E_T(\hat{y}).
\end{equation}
We then define the SimLoss reward as
\begin{equation}
r_{\mathrm{Sim}}(v,\hat{y}) = \cos(z^I,z^T).
\end{equation}
This reward is high when the generated caption is close to the image in the embedding space, and low when the caption fails to preserve image-specific information. We optimize this reward using GRPO~\citep{shao2024deepseekmathpushinglimitsmathematical}. For each image-prompt pair, the policy samples a group of captions $\{\hat{y}_1,\ldots,\hat{y}_G\}$. Each caption receives a reward $r_g = r_{\mathrm{Sim}}(v,\hat{y}_g)$. GRPO estimates relative advantages within the sampled group and updates the policy to increase the likelihood of captions with higher embedding alignment. This allows SimLoss to be used even when the embedding model cannot provide gradients.

\subsection{FeedQuill PPO Baseline}
\label{sec:reward}

We evaluate FeedQuill using PPO with a composite caption-quality reward. For each sampled caption $\hat{y} \sim \pi_\theta(\cdot \mid x,v)$, we compute
\begin{equation}
R(\hat{y},v) =
R_{\mathrm{DC}}(\hat{y},v)
+ R_{\mathrm{CLIP}}(\hat{y},v)
+ R_{\mathrm{CIDEr}}(\hat{y}),
\end{equation}
where $R_{\mathrm{DC}}$ is the judge-computed F1 over the caption's decomposed atomic units, $R_{\mathrm{CLIP}}$ measures global image--text alignment, and $R_{\mathrm{CIDEr}}$ measures similarity to pseudo-references. The first term combines the judge's unit-level precision and recall as $2PR/(P+R)$. This baseline tests whether directly optimizing external caption-quality metrics with PPO can improve fine-grained captioning. In contrast, SimLoss FFT does not use caption references or judge-model rewards. Its supervision comes only from matching the VLM's internal representation to a frozen image embedding.

\section{Baselines}
\label{sec:baselines}

We compare SimLoss against four families of baselines: plain zero-shot captioning, inference-time verification, reward-based RL, and perception-aware policy optimization. Relative to a plain single-pass captioner, these baselines test whether fine-grained captioning can be improved by external verification, reward optimization, or explicit visual perturbation.

\subsection{Plain Captioning Baseline}

The plain baseline uses the base Qwen2.5-VL-7B captioner without multi-stage verification or additional training. It represents the single-pass model before SimLoss adaptation and provides the reference point for measuring whether training improves fine-grained precision, recall, and F1.

\subsection{CapMAS: Multi-Pass Verification Pipeline}

CapMAS~\citep{lee2025robusthyperdetailedimagecaptioning} is a five-stage inference-time pipeline: it samples five captions, merges them, decomposes the result into atomic propositions, visually verifies each proposition, and rewrites the caption from the retained claims. It is a strong verification baseline but requires multiple VLM and LLM calls. We use it to test whether training-time SimLoss can approach the same quality with single-pass inference; full pipeline prompts are in Appendix~\ref{sec:prompts_capmas}.

\subsection{FeedQuill Reward Optimization}

FeedQuill~\citep{ye2025paintingwordselevatingdetailed} evaluates a detailed caption after decomposing it into smaller, atomic information units. A judge scores the decomposed units to obtain unit-level precision and recall, whose harmonic mean provides an F1 reward for PPO. This baseline tests whether fine-grained captioning can be improved by optimizing an external, judge-computed caption-quality reward rather than by direct representation alignment.

\subsection{PAPO: Perception-Aware Policy Optimization}

PAPO~\citep{wang2025perceptionawarepolicyoptimizationmultimodal} adds a perception loss to GRPO by computing the KL divergence between the model's output distribution on the original image and on randomly masked image variants. 
We evaluate PAPO with random masking and additionally extend it with object-aware masking using YOLOv13~\citep{lei2025yolov13realtimeobjectdetection} detections. 

\section{Experiments}
\label{sec:experiments}

\subsection{Setup}

\paragraph{Training data.}
We train on MS~COCO images. For SimLoss, the COCO captions are discarded: FFT uses only images and prompts, with the frozen embedding model providing the adaptation signal. Reward-based baselines use the signals described in Section~\ref{sec:reward}.

\paragraph{Model.}
All trainable methods use 7B-class Qwen vision-language models fine-tuned with LoRA. For SimLoss FFT, the frozen encoder is Qwen3-VL-Embed and the projector maps VLM hidden states into the image-embedding space. The frozen encoder and projector are used only during training and are discarded at inference time. For SimLoss GRPO, Gemini~2~Embed provides a black-box image-text alignment reward.

\paragraph{Evaluation.}
We evaluate on IIW-400 using CLAIR, precision, recall, and F1 as defined in Section~\ref{sec:evaluation_measures}; full judge prompts are in Appendix~\ref{sec:prompts_eval}. We additionally report caption length and matched A100 latency in seconds per image.

\subsection{Main Results}

\begin{table*}[t]
\centering
\small
\begin{tabular}{ll|c|ccc|c}
\toprule
Family & Method/Model & CLAIR & Precision & Recall & F1 & Length (words)  \\
\midrule
\multicolumn{7}{l}{\textit{Inference-time baselines}} \\
Plain & Qwen2.5-VL-7B & 0.842 & 0.7884 & 0.6002 & 0.6815 & $347.55_{\pm 46.82}$ \\
CapMAS & 5-stage pipeline & 0.854 & \underline{0.8467} & \underline{0.6003} & \textbf{0.7025} & $189.71_{\pm 49.03}$ \\
\midrule
\multicolumn{7}{l}{\textit{PAPO baselines}} \\
PAPO & Qwen2.5-7B & 0.843 & 0.7905 & 0.5950 & 0.6790 & $181.86_{\pm 30.47}$ \\ 
PAPO + YOLO & Qwen2.5-7B & 0.842 & 0.7936 & 0.5969 & 0.6813 & $182.41_{\pm 32.71}$ \\ 
\midrule
\multicolumn{7}{l}{\textit{Reward-optimized baselines}} \\
FeedQuill & Qwen2.5-VL-7B & 0.833 & 0.7966 & 0.5967 & 0.6823 & $164.15_{\pm 69.48}$ \\ 
\midrule
\multicolumn{7}{l}{\textit{Ours}} \\
SimLoss FFT & Qwen3-VL-Embed & 0.844 & \textbf{0.8485} & 0.5991 & \underline{0.7023} & \textbf{$114.86_{\pm 13.06}$}  \\
SimLoss GRPO & Gemini 2 Embed & \textbf{0.858} & 0.8227 & \textbf{0.6015} & 0.6949 & $132.87_{\pm 34.02}$ \\
SimLoss-PAPO-YOLO & Qwen2.5-7B & \underline{0.855} & 0.8340 & 0.5970 & 0.6959 & $149.84_{\pm 29.56}$ \\
\bottomrule
\end{tabular}
\caption{Results on IIW-400. SimLoss FFT has the highest precision and nearly matches multi-stage CapMAS in F1 with single-pass inference; SimLoss GRPO has the highest CLAIR and recall. Length is mean $\pm$ standard deviation in words.}
\label{tab:results}
\end{table*}
Table~\ref{tab:results} compares plain zero-shot captioning, the multi-stage CapMAS pipeline, PAPO variants, the PPO-trained FeedQuill baseline, and our SimLoss variants. The strongest methods are CapMAS and SimLoss FFT. CapMAS obtains the highest F1 score, 0.7025, while SimLoss FFT reaches 0.7023, only 0.0002 lower. SimLoss FFT also achieves the highest precision among all methods, 0.8485, slightly exceeding CapMAS at 0.8467. These results support our central claim: image-embedding supervision can adapt a fine-grained captioner without ground-truth or pipeline-generated caption targets while recovering nearly all of CapMAS's F1 benefit in a single pass.

\paragraph{SimLoss FFT gives the best precision with the shortest captions.}
SimLoss FFT improves precision from $0.7884$ to $0.8485$ while cutting mean caption length from $347$ words to $114$, with a standard deviation of $13$ against the baseline's $46$.
This admits a stronger reading than ``shorter is not worse.'' Holding grounded visual content approximately fixed, as the flat recall confirms it is, a caption conveying that content in fewer words is a more efficient textual encoding of the image. 
This is consistent with the objective: it encourages the internal representation to retain the information that identifies the source image, rather than to reproduce a verbose reference or optimize for length-correlated rewards.
In the noisy-channel terms of Section~\ref{sec:formulation}, the model preserves $I(v;\hat{y})$ while spending fewer symbols, trimming generic phrasing rather than visual evidence. 

\paragraph{SimLoss GRPO improves recall but is less precise.}
SimLoss GRPO achieves the highest recall (0.6015) and CLAIR (0.858), suggesting that black-box embedding rewards encourage broader semantic coverage. However, its precision is lower than SimLoss FFT, 0.8227 versus 0.8485, resulting in a lower F1 score of 0.6949. This pattern is consistent with the less direct optimization of sampled text compared with the fully differentiable FFT objective.


\paragraph{Overall comparison.}
CapMAS remains marginally best in F1, but it relies on a multi-stage inference pipeline. SimLoss FFT nearly matches CapMAS in F1, surpasses it in precision, produces the shortest and most stable captions, and keeps inference single-pass. SimLoss GRPO is useful as a black-box reward variant and achieves the best recall, but it is less precise than FFT. An extended analysis of why each method exhibits this behavior is provided in Appendix~\ref{sec:analysis}, and qualitative examples examining where and why these differences arise are provided in Appendix~\ref{sec:qualitative}.

\subsection{Latency and Quality-Latency Tradeoff}
\begin{table}[t]
\centering
\small
\begin{tabular}{lccc}
\toprule
Method & F1 & sec/img & Speedup \\
\midrule
CapMAS & \textbf{0.7025} & 115.31 & 1.0$\times$ \\
Plain Qwen2.5-VL-7B & 0.6815 & 8.66 & 13.3$\times$ \\
FeedQuill & 0.6823 & 8.97 & 12.9$\times$ \\
PAPO + YOLO & 0.6813 & 7.30 & 15.8$\times$ \\
SimLoss GRPO (ours) & 0.6949 & \underline{6.58} & \underline{17.5}$\times$ \\
SimLoss-PAPO-YOLO (ours) & 0.6959 & 7.21 & 16.0$\times$ \\
SimLoss FFT (ours) & \underline{0.7023} & \textbf{5.77} & \textbf{20.0$\times$} \\
\bottomrule
\end{tabular}
\caption{End-to-end inference latency on A100. SimLoss FFT nearly matches CapMAS in F1 while being about 20$\times$ faster. The gap primarily reflects single-pass versus multi-stage inference; differing output lengths also affect wall-clock time.}
\label{tab:latency}
\end{table}

Table~\ref{tab:latency} reports the quality--latency tradeoff. CapMAS obtains the highest F1, but requires 115.31 seconds per image. SimLoss FFT reaches nearly the same F1 at 5.77 seconds per image, yielding a 20.0$\times$ measured speedup. Unlike CapMAS, the deployed model makes one VLM pass because the frozen encoder and projector are training-only; SimLoss FFT's shorter output can also reduce decoding time.



\subsection{SimLoss FFT vs.\ SimLoss GRPO}

SimLoss FFT and SimLoss GRPO expose complementary precision--recall behavior. FFT produces the strongest precision and nearly the best F1, suggesting that differentiable representation alignment is effective for grounded caption generation. GRPO produces the strongest recall, suggesting that black-box embedding rewards can encourage broader visual coverage. However, GRPO's lower precision shows that embedding-reward optimization alone can admit plausible but less grounded details. In this setting, the differentiable FFT objective provides a better operating point when both grounding and inference efficiency are important.

\subsection{Qualitative Analysis and Outlook}
\label{sec:qualitative_summary}

\paragraph{Qualitative summary.}
Across eight IIW-400 examples, SimLoss more consistently describes scene depth, paired objects, visible text, and fine-grained materials while using less repetition. In two matched-coverage cases, SimLoss FFT answers the same questions in roughly one third as many words as the plain or CapMAS captions. The trend is not universal: both SimLoss variants mistake an abstract driftwood moose for a deer, showing that greater specificity can amplify fine-category errors. Full captions and per-theme comparisons appear in Appendices~\ref{sec:qualitative} and~\ref{sec:qual_length}.

\paragraph{Discussion and future work.}
SimLoss's central benefit is adaptation without ground-truth or pipeline-generated captions. This replaces dependence on caption targets with dependence on a frozen embedding teacher: its biases and visual granularity determine which distinctions are rewarded. Future work should compare or ensemble teachers across domains, combine FFT's precision with GRPO's recall, and explore confidence calibration or lightweight verification that preserves the single-pass deployment path. Further quantitative discussion appears in Appendix~\ref{sec:analysis}.

\section{Conclusion}

SimLoss adapts a VLM without ground-truth caption targets by aligning its projected hidden-state representation with frozen image embeddings, encouraging the model to preserve discriminative visual information before text generation occurs.
On IIW-400, SimLoss FFT achieves the highest precision among all evaluated methods, edging out CapMAS and nearly matching its F1 while retaining single-pass inference. SimLoss GRPO achieves the highest recall, evidence that embedding-space rewards can broaden visual coverage even without access to embedding-model gradients.

The remaining gap to CapMAS suggests that explicit verification is still useful for precision-recall balancing. A natural next direction is to combine SimLoss-trained single-pass captioners with lightweight factuality checks, aiming to preserve the latency benefits of single-pass generation while further improving factual consistency.


\bibliography{references}


\clearpage
\newpage
\appendix

\section{Analysis}
\label{sec:analysis}

\paragraph{Why does SimLoss FFT improve precision?}
SimLoss FFT directly aligns the VLM's continuous hidden-state representation with a frozen image embedding. This creates a dense training signal before text generation occurs. Unlike caption imitation, SimLoss does not force the model to copy the wording or omissions of a pseudo-reference. Unlike GRPO, it does not rely on high-variance updates from sampled text. The resulting model is encouraged to preserve image-specific evidence in its internal representation, which likely explains its strong precision and near-CapMAS F1.

\paragraph{Why does SimLoss GRPO improve recall but not precision?}
SimLoss GRPO uses a black-box embedding model to reward generated captions whose text embeddings align with the image embedding. This reward can encourage broader visual coverage, which is reflected in the highest recall score of 0.6015. However, because the reward is applied after discrete generation, optimization is less direct than FFT. The model can increase embedding similarity by adding more visually plausible content, but this can reduce precision. This explains why SimLoss GRPO improves recall while lagging SimLoss FFT in precision and F1.

\paragraph{Why is CapMAS still marginally best in F1?}
CapMAS explicitly decomposes captions into propositions, verifies each proposition against the image, and rewrites the final caption using verified claims. This inference-time verification is well suited to controlling false positives, which gives CapMAS a marginal F1 advantage. However, this advantage is extremely small in our results: 0.7025 for CapMAS vs.\ 0.7023 for SimLoss FFT. Given the latency gap between a multi-stage pipeline and a single-pass model, SimLoss FFT occupies a more favorable quality-latency point.

\paragraph{Why do reward-style baselines underperform?}
FeedQuill depends on a judge-model reward optimized with PPO, but the reward signal is noisy and indirect for fine-grained visual recall. PAPO introduces visual sensitivity through masking, but its gains depend strongly on the model variant and masking strategy. Object-aware YOLO masking can further introduce noise when detections are incomplete or when masked objects remove contextual information. In contrast, SimLoss FFT provides a direct differentiable alignment objective between the model's internal visual representation and the frozen image embedding.

\paragraph{LLM-judge supervision is a fragile signal.}
A deeper limitation links the two strongest non-SimLoss approaches. CapMAS verifies each proposition with a multimodal judge, and FeedQuill rewards generations with a judge model during PPO; both therefore inherit the reliability of an LLM judge. Recent work shows that LLM judges suffer from a systematic agreeableness bias: they accept valid outputs almost perfectly (a true-positive rate near 96\%) but flag invalid ones poorly (a true-negative rate below 25\%), and naive judge ensembling such as majority voting does not remove this bias~\citep{jain2025consensusmitigatingagreeablenessbias}. A judge that rarely flags unsupported content is a weak factuality signal whether it is used to filter propositions at inference time or to reward captions during training, since hallucinated but plausible details are exactly the case it misses. SimLoss avoids this dependence: its supervision is a frozen image embedding rather than a judge verdict, so it does not require an external model to reliably detect what is and is not grounded in the image.

\paragraph{Length alone does not explain quality.}
SimLoss FFT also has the lowest mean and standard deviation of generated caption length among the evaluated methods. This suggests that it achieves strong precision and F1 without producing unusually long or highly variable captions. In contrast, the plain baseline has a much larger mean length, yet substantially lower precision and F1. Fine-grained captioning is therefore not simply a matter of generating longer descriptions; the details must be visually grounded. Qualitative examples illustrating this point are provided in Appendix~\ref{sec:qual_length}.

\paragraph{Conciseness as efficient visual encoding.}
The length results admit a stronger reading than ``shorter is not worse.'' SimLoss FFT attains the highest precision and near-best F1 with the shortest captions, averaging 114 words against 347 for the plain baseline and 189 for the verified CapMAS pipeline. Holding the grounded visual content roughly fixed, a caption that conveys it in fewer words is a more efficient textual encoding of the image: the same information, written more economically. This is the spirit of lossless compression, where the aim is to shorten a description without discarding the information it carries. SimLoss does not guarantee losslessness, but its objective pulls in this direction, since it rewards the internal representation for retaining the information needed to identify the source image rather than for reproducing a verbose reference or inflating a length-correlated reward. In the noisy-channel terms of Section~\ref{sec:formulation}, the model is encouraged to preserve the mutual information $I(\text{image};\text{caption})$ while spending fewer symbols, trimming generic or redundant phrasing rather than the visual evidence itself. Conciseness here is thus a byproduct of optimizing for information content: the most efficient way to keep a caption discriminative is to fill it with grounded detail and little else.

\section{Qualitative Analysis}
\label{sec:qualitative}

\begin{figure}[!ht]
    \centering
    \includegraphics[width=0.45\linewidth]{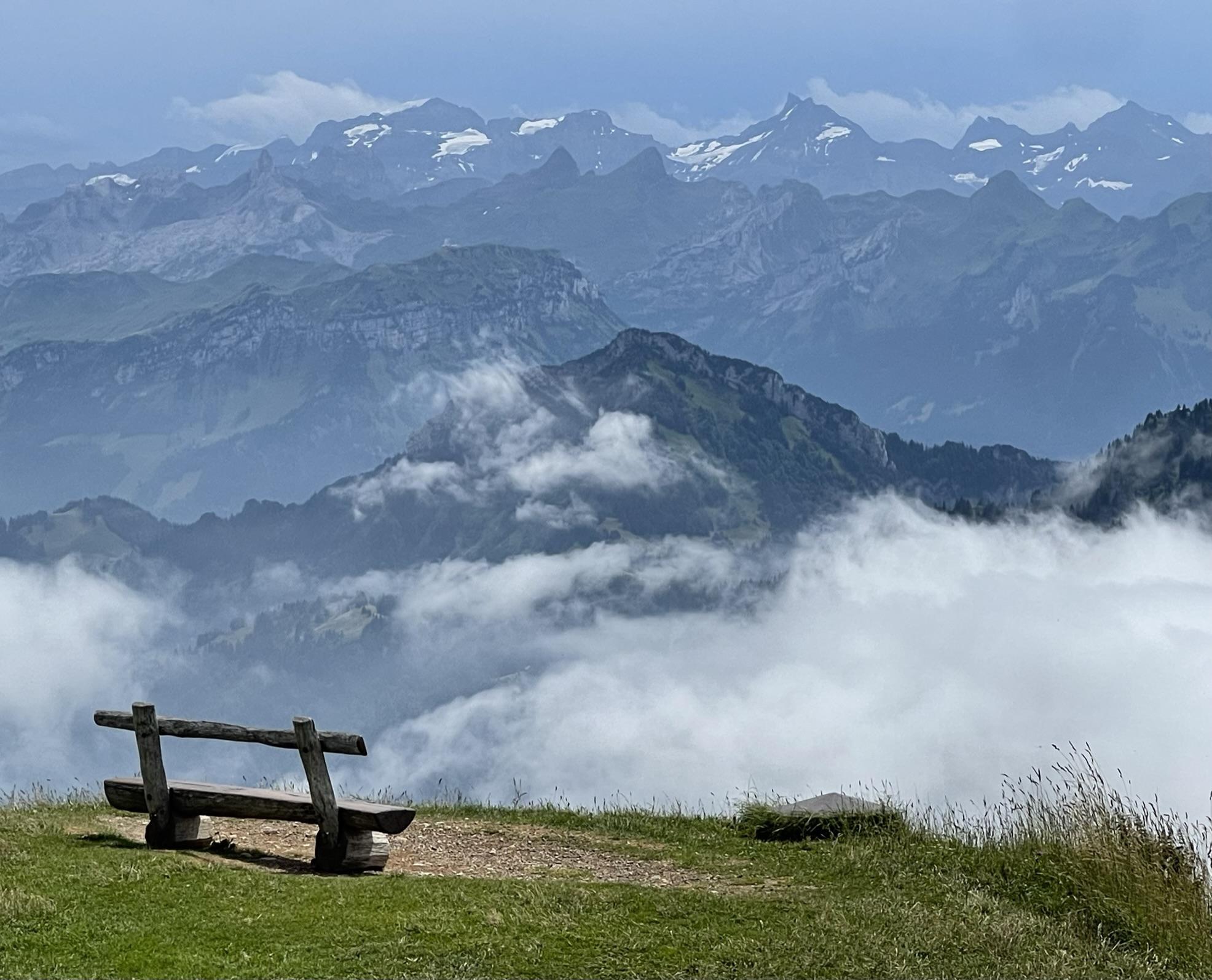}\\[4pt]
    {\scriptsize\texttt{aar\_test\_04859}}

    \parbox{0.9\linewidth}{\footnotesize \textbf{SimLoss FFT, 122 words.} ``This image captures a serene and majestic mountain landscape. In the foreground, there is a wooden bench placed on a grassy area, inviting viewers to sit and enjoy the view. The bench is simple in design, with a backrest and seat made of natural wood, blending harmoniously with the natural surroundings.

The middle ground is dominated by a sea of clouds that blanket the lower slopes of the mountains, creating a misty and ethereal atmosphere. The clouds appear thick and low-lying, partially obscuring the lower parts of the mountains and adding a sense of depth and mystery to the scene.

In the background, the mountains rise majestically, their peaks sharp and jagged, with some areas covered in snow, indicating higher altitudes. The''}

    \caption{\textbf{SimLoss FFT caption on \texttt{aar\_test\_04859}}}
    \label{fig:simloss_fft_a}
\end{figure}

\begin{figure}[!ht]
    \centering
    \includegraphics[width=0.45\linewidth]{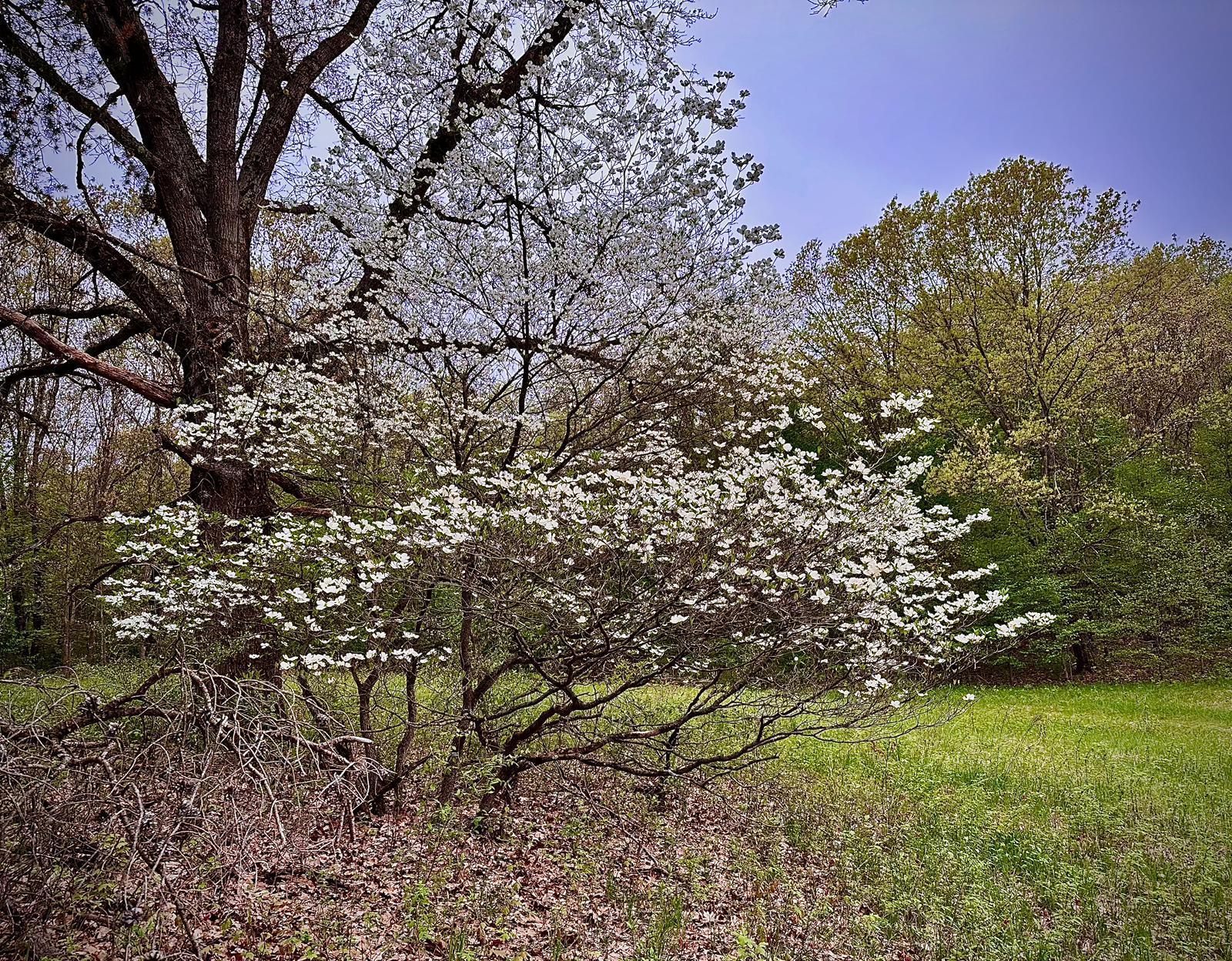}\\[4pt]
    {\scriptsize\texttt{aar\_test\_04715}}

    \parbox{0.9\linewidth}{\footnotesize \textbf{SimLoss FFT, 132 words.} ``This image captures a serene forest scene during what appears to be the spring season, as indicated by the abundance of white flowers blooming on the trees. The foreground features a large tree with a thick trunk and numerous branches, heavily laden with white blossoms. The branches extend outward, creating a dense canopy of flowers that dominate the view. The ground is covered with a mix of green grass and fallen leaves, suggesting a transition between seasons.

In the background, there are other trees with a mix of green foliage and some bare branches, indicating that they might not have fully bloomed yet or are in different stages of growth. The sky above is clear and blue, adding to the overall tranquility of the scene. The lighting suggests it is daytime, possibly''}

    \caption{\textbf{SimLoss FFT caption on \texttt{aar\_test\_04715}}}
    \label{fig:simloss_fft_f}
\end{figure}

\begin{figure}[!ht]
    \centering
    \includegraphics[width=0.45\linewidth]{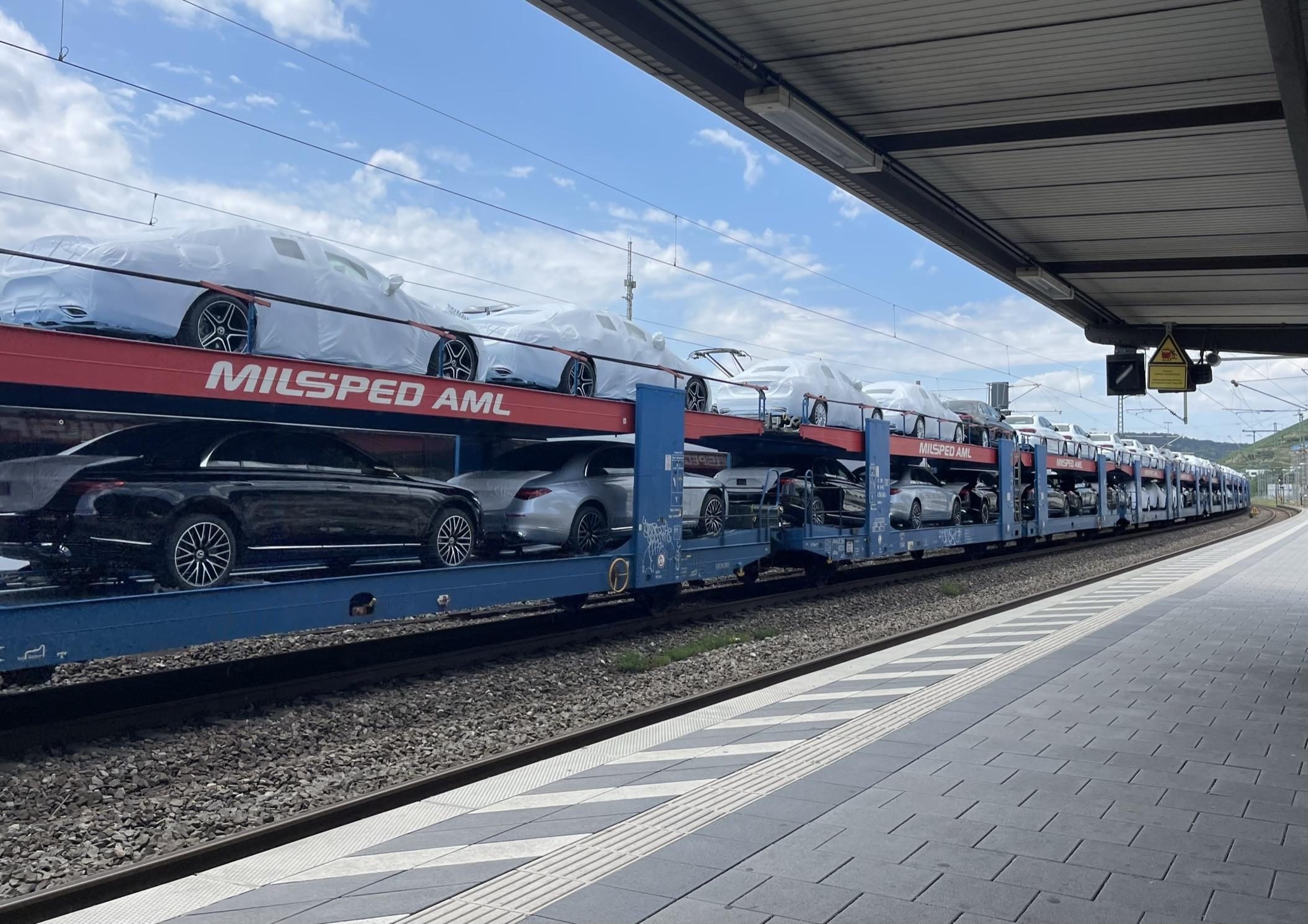}\\[4pt]
    {\scriptsize\texttt{aar\_test\_04615}}

    \parbox{0.9\linewidth}{\footnotesize \textbf{SimLoss FFT, 108 words.} ``The image shows a train platform with a series of car carriers loaded with luxury cars. The train is labeled ``MILSPED AML,'' which likely stands for a company that specializes in the transportation of vehicles by rail. The cars are neatly arranged on the carriers, with some covered in protective plastic to prevent damage during transit. The platform has a modern design with a checkered pattern on the edge, and the sky above is clear with a few scattered clouds. The train is stationary, and the platform appears to be empty except for the cars. The overall scene suggests a professional and organized process for transporting high-value vehicles.''}

    \caption{\textbf{SimLoss FFT caption on \texttt{aar\_test\_04615}}}
    \label{fig:simloss_fft_c}
\end{figure}

\begin{figure}[!ht]
    \centering
    \includegraphics[width=0.45\linewidth]{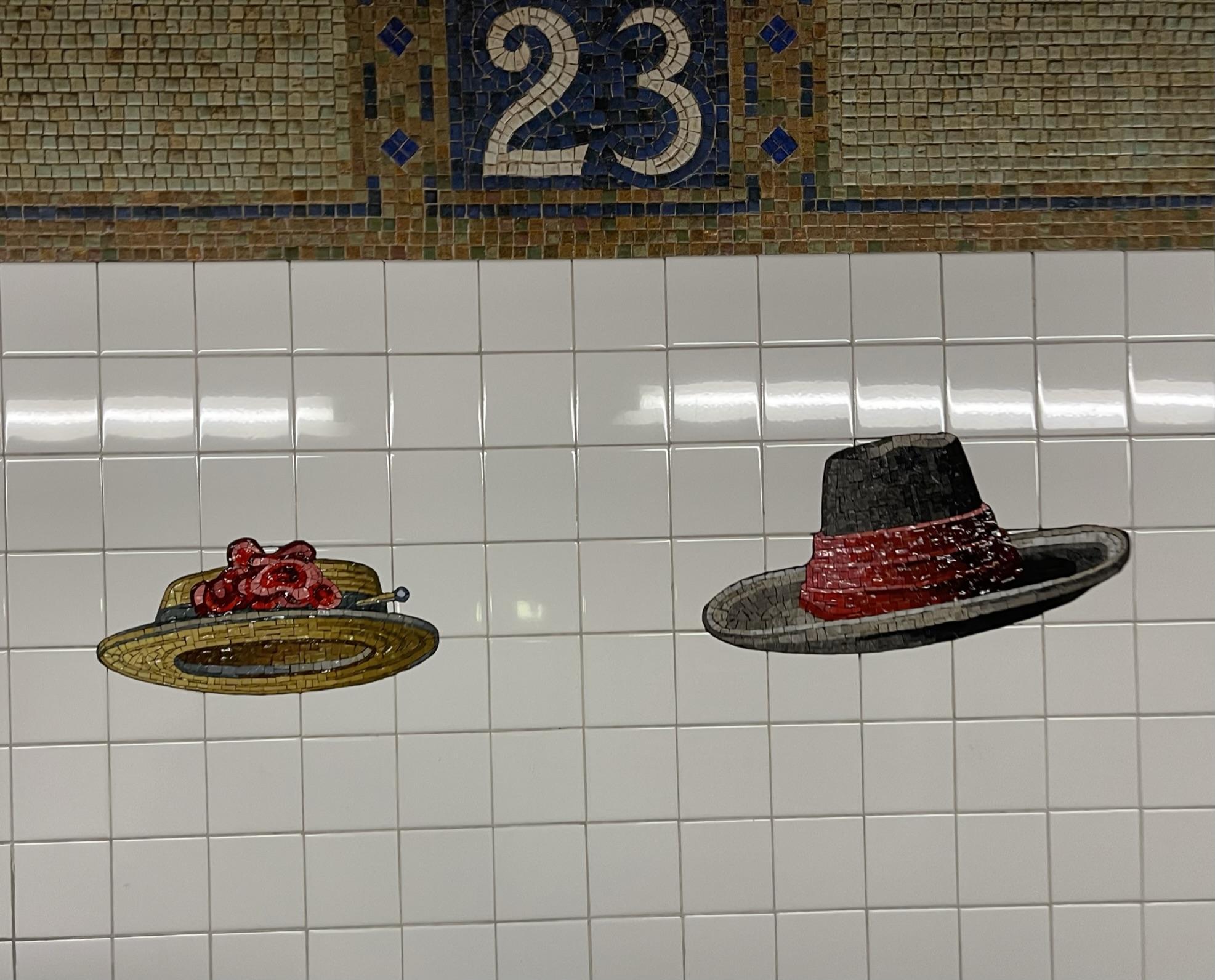}\\[4pt]
    {\scriptsize\texttt{aar\_test\_04667}}

    \parbox{0.9\linewidth}{\footnotesize \textbf{SimLoss FFT, 133 words.} ``The image shows a section of a tiled wall with two mosaic designs of hats. The tiles above the hats feature a number ``23'' in a mosaic pattern, with blue and white tiles forming the digits. Below the number, there is a row of smaller, multicolored tiles that create a border around the number.

The two hats are depicted on separate plates. The hat on the left plate is yellow with a red band and appears to have a red object, possibly a flower or a decorative item, placed on top. The hat on the right plate is black with a red band and also has a red object on top, which looks similar to the one on the yellow hat. Both hats are detailed with mosaic patterns that give them texture and''}

    \caption{\textbf{SimLoss FFT caption on \texttt{aar\_test\_04667}}}
    \label{fig:simloss_fft_b}
\end{figure}

\begin{figure}[!ht]
    \centering
    \includegraphics[width=0.45\linewidth]{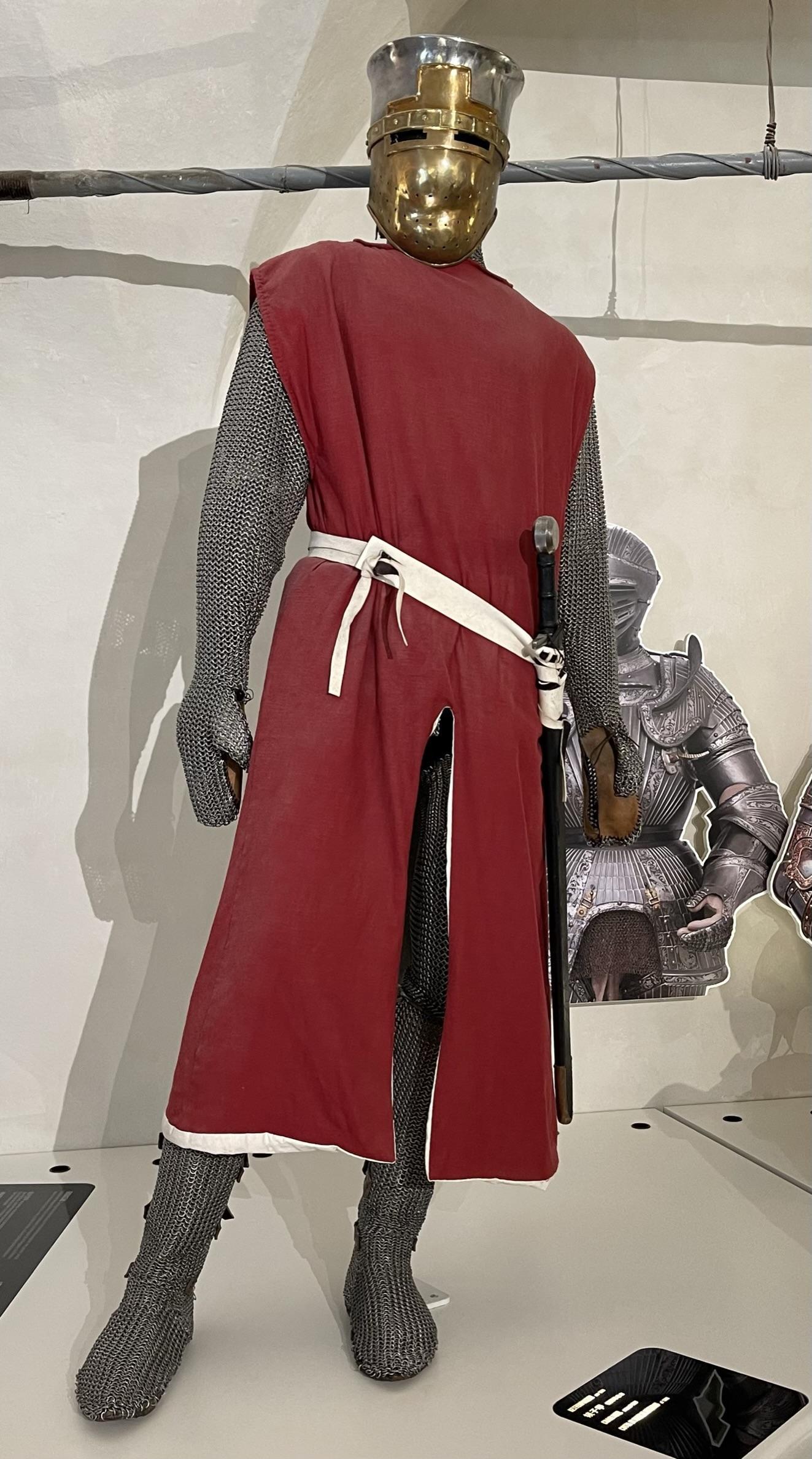}\\[4pt]
    {\scriptsize\texttt{aar\_test\_04696}}

    \parbox{0.9\linewidth}{\footnotesize \textbf{SimLoss FFT, 112 words.} ``The image depicts a mannequin dressed in medieval armor and attire. The mannequin is wearing a chainmail hauberk, which is a type of armor made of interlinked metal rings. Over the hauberk, there is a red surcoat, which is a loose-fitting garment often worn over armor for protection and visibility. The surcoat is secured with a white sash tied around the waist. The mannequin also wears chainmail gauntlets on the arms and chainmail boots that cover the legs up to the knees. A helmet with a visor is worn on the head, providing head protection. The mannequin is holding a sword in its right hand, which is sheathed at the hilt. In''}

    \caption{\textbf{SimLoss FFT caption on \texttt{aar\_test\_04696}}}
    \label{fig:simloss_fft_e}
\end{figure}

\begin{figure}[!ht]
    \centering
    \includegraphics[width=0.45\linewidth]{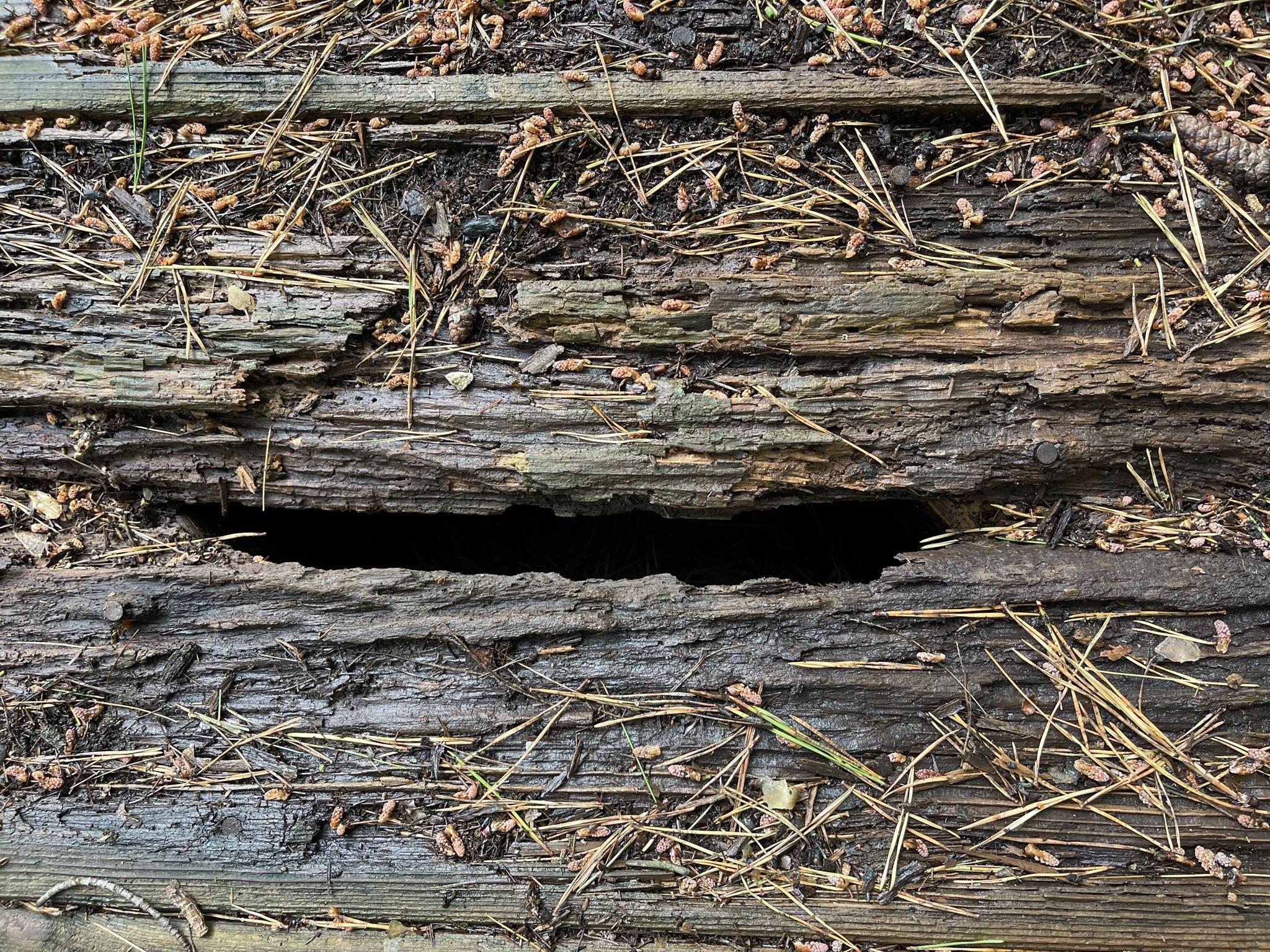}\\[4pt]
    {\scriptsize\texttt{aar\_test\_04657}}

    \parbox{0.9\linewidth}{\footnotesize \textbf{SimLoss FFT, 113 words.} ``The image shows a close-up view of a decaying log. The log is covered with a layer of pine needles and small pine cones, indicating it is part of a forest floor. The wood appears weathered and broken, with a dark, hollow center that suggests it has been hollowed out by decay or possibly by an animal. The texture of the wood is rough and uneven, with cracks and splits visible throughout. The pine needles and cones are scattered around the log, adding to the natural debris on the forest floor. The overall appearance of the log gives a sense of age and decomposition, typical of a fallen tree in a forest environment.''}

    \caption{\textbf{SimLoss FFT caption on \texttt{aar\_test\_04657}}}
    \label{fig:simloss_fft_d}
\end{figure}

\begin{figure}[!ht]
    \centering
    \includegraphics[width=0.45\linewidth]{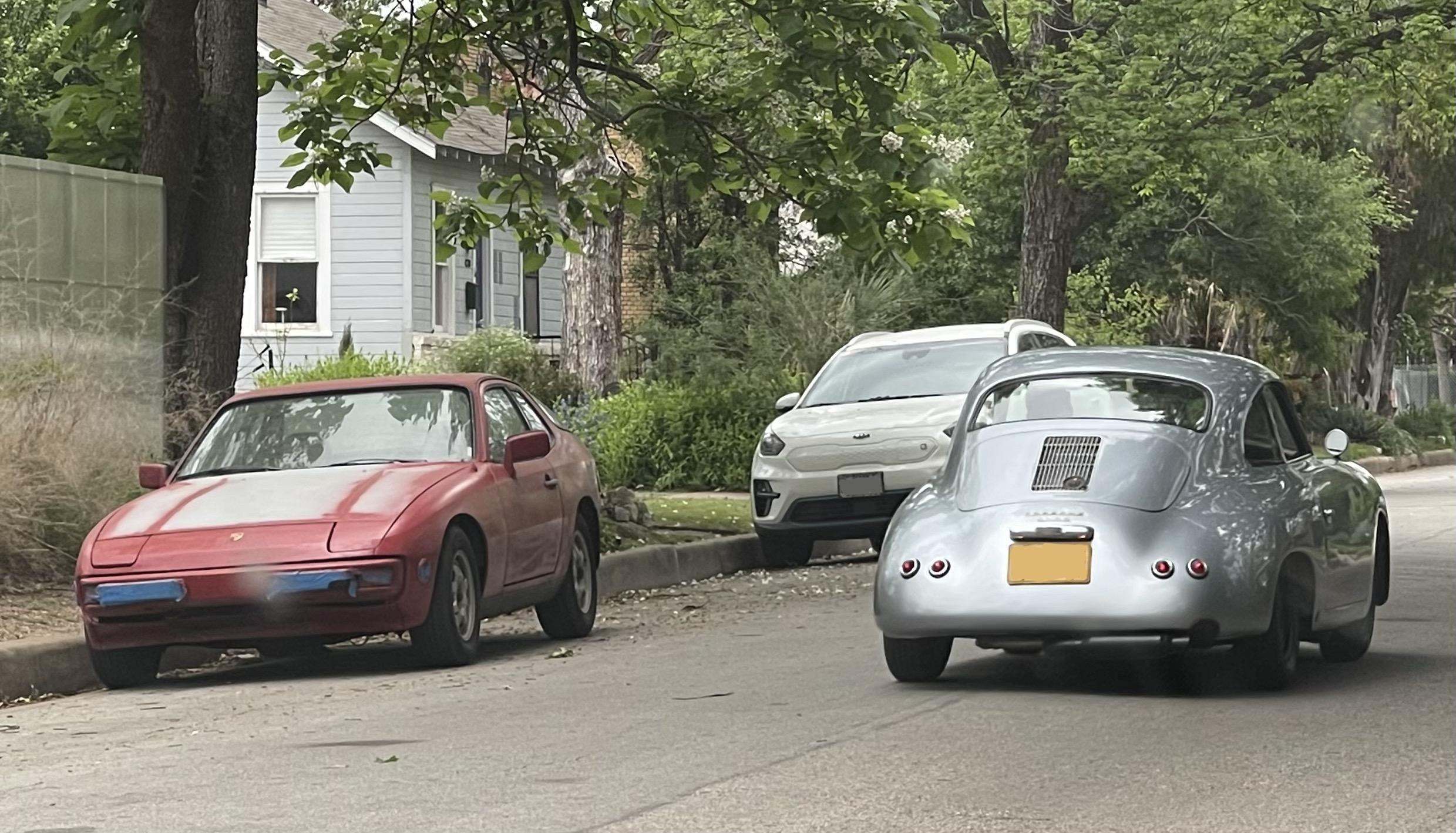}\\[4pt]
    {\scriptsize\texttt{aar\_test\_04720}}

    \parbox{0.9\linewidth}{\footnotesize \textbf{SimLoss FFT, 130 words.} ``The image shows three cars parked on the side of a street in a residential area. The car on the left is a red Porsche, which appears to be an older model with a classic design. The middle car is a white SUV, which looks more modern and is parked behind the Porsche. The car on the right is a silver vintage car, possibly a classic Porsche as well, given its shape and design. The cars are parked in front of a house with a light-colored exterior and a tree in the background. The street is lined with trees and bushes, and there is a fence partially visible on the left side of the image. The overall scene suggests a quiet neighborhood with a mix of classic and modern vehicles.''}

    \caption{\textbf{SimLoss FFT caption on \texttt{aar\_test\_04720}}}
    \label{fig:simloss_fft_g}
\end{figure}

\begin{figure}[!ht]
    \centering
    \includegraphics[width=0.45\linewidth]{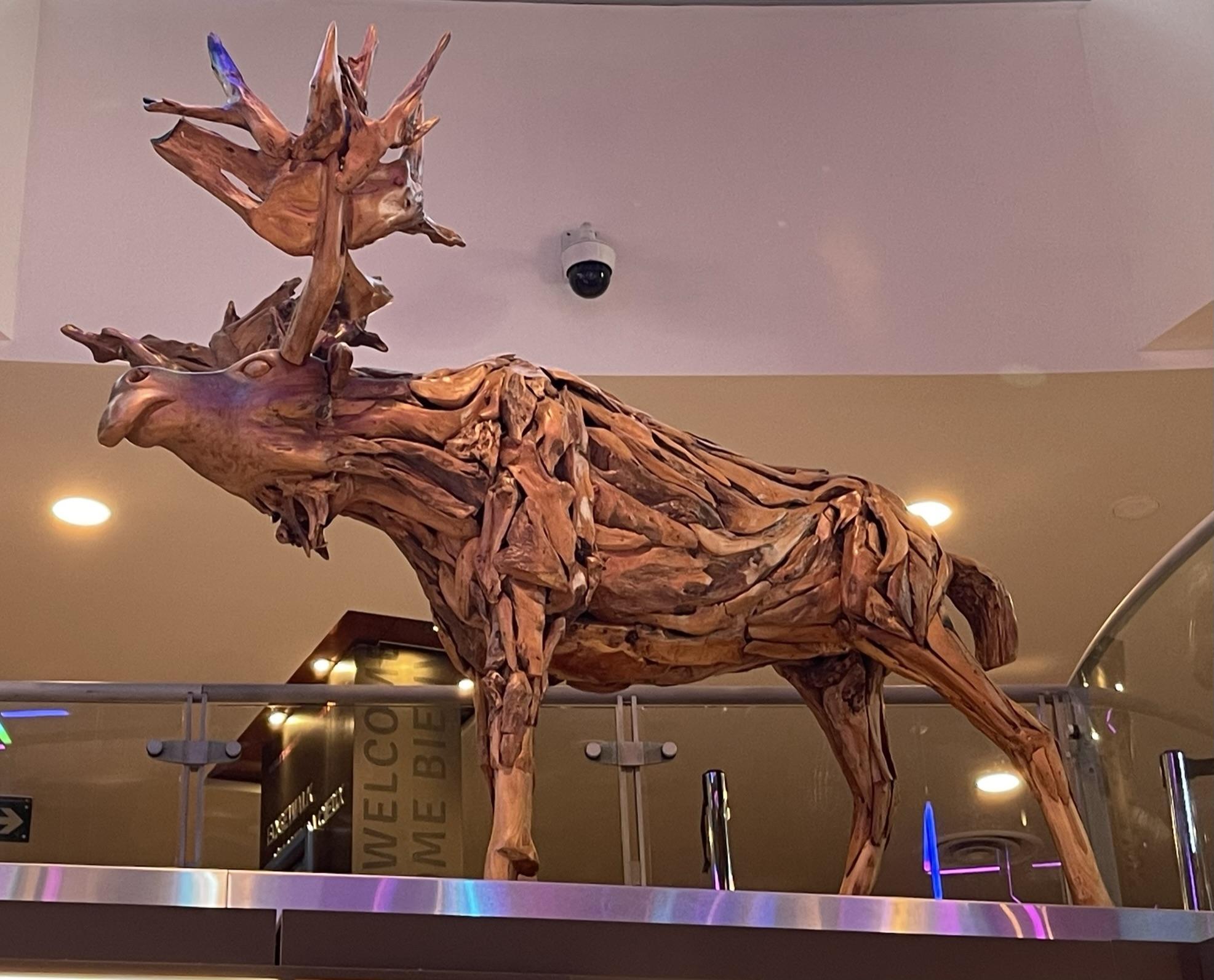}\\[4pt]
    {\scriptsize\texttt{aar\_test\_04873}}

    \parbox{0.9\linewidth}{\footnotesize \textbf{SimLoss FFT, 121 words.} ``The image depicts a large, intricately carved wooden sculpture of a deer. The sculpture is highly detailed, with the antlers and body of the deer crafted from wood that has been carefully shaped to mimic the natural form of the animal. The antlers are particularly elaborate, with multiple branches and points, giving the sculpture a dynamic and lifelike appearance. The deer's head is turned slightly to the side, and its mouth appears to be open, as if it is mid-roar or mid-breathe.

The sculpture is mounted on a platform, which is part of an indoor setting, likely a museum or gallery, as indicated by the ``WELCOME'' sign in the background. The lighting in the room is warm, and there are lights''}

    \caption{\textbf{SimLoss FFT caption on \texttt{aar\_test\_04873}}}
    \label{fig:simloss_fft_h}
\end{figure}

    \noindent We examine captions from all methods on the images shown in Figures~\ref{fig:simloss_fft_a}--\ref{fig:simloss_fft_h} to understand where and why the quantitative gaps arise. Four themes emerge from this comparison.

\paragraph{Scene structure.}
Most baselines describe images as an unordered list of objects. For Figure~\ref{fig:simloss_fft_a}, Plain VLM, CapMAS, and FeedQuill/PPO all mention a bench, mountains, and clouds, but treat them as co-equal elements rather than things at different depths. Plain VLM goes further and calls the sky ``clear, with a light blue hue'' - a straightforward error given the prominent cloud bank in the image. PAPO and PAPO+YOLO do better but still treat the cloud layer as incidental atmosphere rather than a distinct mid-ground feature. SimLoss FFT and SimLoss GRPO both explicitly organize the scene into three layers: bench in the foreground, a ``sea of clouds that blanket the lower slopes'' in the middle ground, and the snow-capped peaks in the background. This layered structure directly covers more spatial and relational questions in the evaluation set, which is the main driver of the coverage gain on this image.

Figure~\ref{fig:simloss_fft_f} shows a related difference. PAPO and PAPO+YOLO correctly name the flowers as dogwood---an improvement over Plain VLM and CapMAS, which say only ``white blossoms''---but neither mentions the ground. SimLoss FFT records the visible mix of green grass and fallen leaves, and SimLoss GRPO similarly describes ``green grass and scattered fallen leaves'' beneath the tree. No baseline captures this ground-level evidence.

\paragraph{Reading visible text.}
For Figure~\ref{fig:simloss_fft_c}, Plain VLM and CapMAS do not mention the ``MILSPED AML'' label on the railcar at all. PAPO and PAPO+YOLO attempt to read it but get it wrong (``MILSPEED AML''), and PAPO+YOLO invents an expansion---``Armored Military Locomotive''---with no basis in the image. FeedQuill/PPO and both SimLoss variants transcribe ``MILSPED AML'' correctly; SimLoss GRPO additionally records the platform's tactile paving strip, a visible detail the baselines omit.

For Figure~\ref{fig:simloss_fft_b}, the coverage gap is particularly large: our methods answer nine more questions than Plain VLM (32 vs.\ 23). Plain VLM and PAPO describe the left hat and leave the right one vague. CapMAS and FeedQuill/PPO acknowledge both hats but give them unequal treatment. SimLoss FFT describes each hat at the same level of detail - noting that both have a red band and a red decorative object on top - and additionally identifies ``a row of smaller, multicolored tiles that create a border'' between the ``23'' medallion and the hat panels. SimLoss GRPO similarly gives symmetric descriptions of both hats, specifying ``yellow hat with red flowers'' on the left and ``black hat with a red band'' on the right.

\paragraph{Specificity of object description.}
For Figure~\ref{fig:simloss_fft_e}, PAPO+YOLO gets the period right but uses only generic terms. SimLoss FFT names the garments precisely: ``chainmail hauberk,'' ``red surcoat secured with a white sash,'' ``chainmail gauntlets,'' and ``chainmail boots.'' SimLoss GRPO is similarly specific, correctly identifying the helmet type and the white belt cinching the surcoat at the waist. No baseline uses terminology at this level of specificity.

For Figure~\ref{fig:simloss_fft_d}, PAPO and PAPO+YOLO describe the debris on the log as ``moss and small twigs,'' neither of which is visible in the image. Plain VLM hedges with ``light-colored pieces that could be bits of bark or other organic debris.'' SimLoss FFT identifies the material correctly as ``pine needles and small pine cones,'' and SimLoss GRPO goes further, noting that ``some of the pine needles are dry and light brown, while others are still green'' - a level of granularity that supports questions about the state of the debris and the surrounding ecosystem.

For Figure~\ref{fig:simloss_fft_g}, SimLoss FFT and SimLoss GRPO venture a brand-level identification of the red sports car as a Porsche and suggest that the silver vintage car is another Porsche. This extra specificity can improve coverage when correct, but it also creates fine-category risk. Figure~\ref{fig:simloss_fft_h} exposes that tradeoff: our methods describe the antlers and body of the abstract driftwood sculpture in detail but call the moose a deer, whereas the more conservative baselines avoid this error.

\paragraph{Hallucinated inferences.}
Several baselines add claims that go beyond what the image shows. For Figure~\ref{fig:simloss_fft_a}, CapMAS describes the location as ``likely a popular hiking or nature-watching spot.'' For Figure~\ref{fig:simloss_fft_c}, it asserts the platform is ``well-maintained, with no visible debris or litter.'' PAPO+YOLO's fabricated AML expansion is another example. These plausible but unverifiable claims coincide with precision losses. On Figure~\ref{fig:simloss_fft_a}, both SimLoss variants instead describe the prominent cloud layer that Plain VLM misses, and SimLoss FFT attains 100\% measured precision versus 96.3\% for Plain VLM and 85.2\% for CapMAS. Other examples above show that SimLoss can still over-specify, so the improvement is a tendency rather than a guarantee.

\begin{figure*}[!ht]
    \centering
    \captionsetup[subfigure]{font=footnotesize}
    \begin{minipage}[t]{0.27\textwidth}
        \vspace{0pt}
        \centering
        \includegraphics[width=\linewidth]{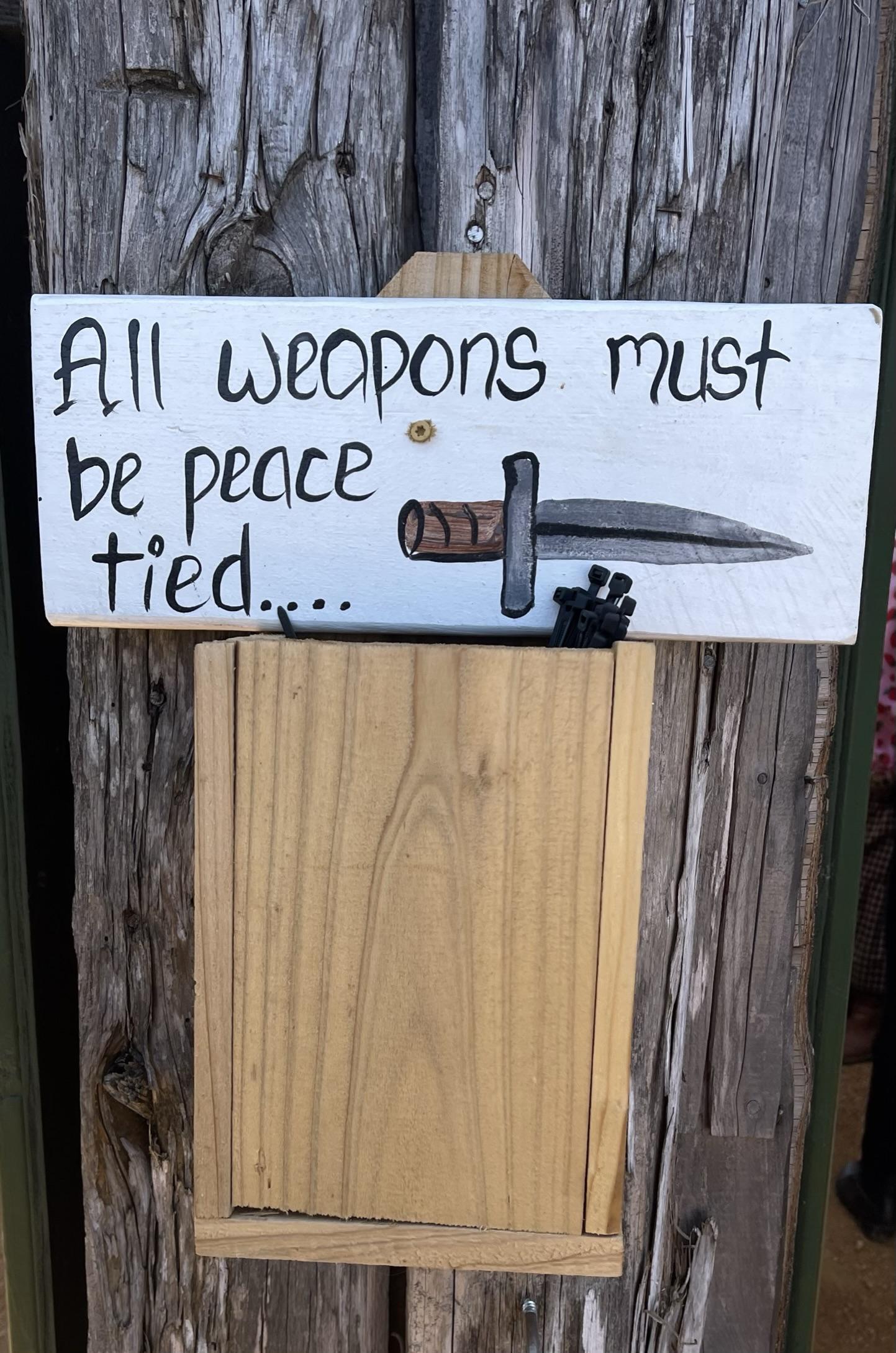}\\[2pt]
        {\scriptsize\texttt{aar\_test\_04779}}
    \end{minipage}\hfill
    \begin{minipage}[t]{0.70\textwidth}
        \vspace{0pt}
        \begin{subfigure}{\linewidth}
            \caption{\textbf{Plain VLM, 340 words.} ``The image depicts a wooden sign mounted on a weathered wooden post. The sign is rectangular and appears to be handcrafted, with a white background and black text. The text on the sign reads: `All weapons must be peace tied....' The word `peace' is emphasized by being written in a larger, bolder font compared to the rest of the text. To the right of the text, there is a simple illustration of a knife. The knife is depicted with a wooden handle and a metallic blade. The illustration is minimalistic, with the handle being brown and the blade being gray. The knife is positioned horizontally, with the blade pointing to the right. Below the sign, there is a wooden box or container. The box is rectangular and made of light-colored wood, possibly pine or another softwood. The box appears to be empty and is mounted on the same wooden post as the sign. The box is positioned directly beneath the sign, suggesting that it might be used to store or organize items related to the sign's message. The background of the image shows the wooden post, which is weathered and has a rough texture. The wood appears to be old and has a natural, rustic appearance. The post is vertical and supports both the sign and the box. The overall setting suggests a rustic or outdoor environment, possibly a campsite, a cabin, or a similar location where the sign is meant to be visible and easily understood by visitors or residents. The sign's message, `All weapons must be peace tied....', implies a rule or guideline for the safekeeping of weapons, possibly indicating that weapons should be secured or tied in a way that ensures they are not accessible or dangerous. In summary, the image shows a handcrafted wooden sign with a message about the safekeeping of weapons, accompanied by a simple illustration of a knife and a wooden box mounted on a weathered wooden post. The setting appears to be rustic and outdoor, likely in a campsite or similar environment.''}
        \end{subfigure}

        \medskip
        \begin{subfigure}{\linewidth}
            \caption{\textbf{CapMAS, 133 words.} ``The image depicts a rustic, weathered wooden post supporting a rectangular white sign with black text and a black-and-white illustration of a knife. The sign reads, `All weapons must be peace tied....' The illustration shows a knife with a wooden handle and a metal blade. Below the sign, a light-colored wooden box or container is mounted on the same post, with a flat top and a front panel that appears slightly recessed. The box has a simple design and is likely used for storing or displaying items related to the sign's message. The wooden post is old and weathered, with visible signs of aging such as cracks, knots, and discoloration. The post is dark brown, indicating exposure to the elements for a long time. The overall scene suggests a rustic or outdoor setting.''}
        \end{subfigure}

        \medskip
        \begin{subfigure}{\linewidth}
            \caption{\textbf{SimLoss FFT, 123 words.} ``The image shows a wooden sign attached to a wooden door or panel. The sign is white with black text and features an illustration of a knife. The text on the sign reads, `All weapons must be peace tied....' The knife depicted on the sign has a wooden handle and a metallic blade. Below the sign, there is a piece of wood that appears to be part of the door or panel, and it is secured with some black zip ties. The background consists of weathered wooden planks, giving the sign a rustic appearance. The overall tone of the sign suggests a humorous or ironic message about the use of weapons, possibly indicating a playful or lighthearted context rather than a serious one.''}
        \end{subfigure}
    \end{minipage}
    \caption{\textbf{Caption length versus quality on \texttt{aar\_test\_04779}.} All three methods answer the same 20 of 23 questions, but at very different lengths and precision. (a)~Plain VLM and (b)~CapMAS repeat and speculate more; (c)~SimLoss FFT is shorter and less repetitive, though it still interprets the sign's tone.}
    \label{fig:qual_length_a}
\end{figure*}

\section{Caption Length versus Quality}
\label{sec:qual_length}

\begin{figure*}[!ht]
    \centering
    \captionsetup[subfigure]{font=footnotesize}
    \begin{minipage}[t]{0.27\textwidth}
        \vspace{0pt}
        \centering
        \includegraphics[width=\linewidth]{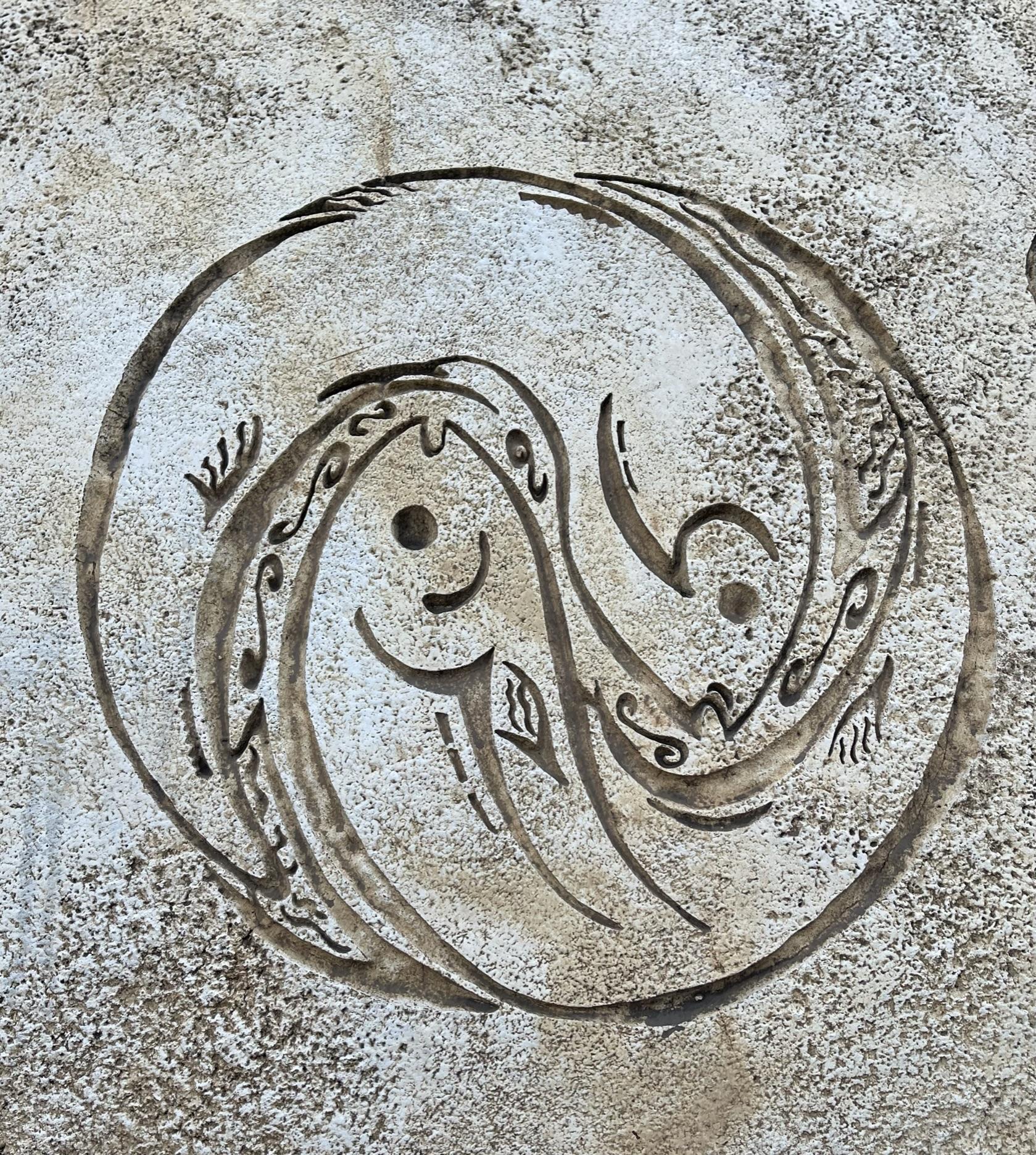}\\[2pt]
        {\scriptsize\texttt{aar\_test\_04629}}
    \end{minipage}\hfill
    \begin{minipage}[t]{0.70\textwidth}
        \vspace{0pt}
        \begin{subfigure}{\linewidth}
            \caption{\textbf{Plain VLM, 395 words.} ``The image depicts a circular stone carving or engraving on a textured surface. The carving is intricate and appears to be a stylized representation of a face or a symbol. The face is composed of several elements that are artistically integrated into the circular design. 1.~Circular Outline: The entire design is contained within a circular boundary, which is the most prominent feature. The circle is not perfectly round but has a slightly irregular shape, giving it a naturalistic feel. 2.~Central Face: At the center of the circle, there is a stylized face. The face is composed of several elements: Eyes: Two small, round eyes are positioned symmetrically on either side of the face. The eyes are simple, with small dots representing the pupils. Nose: A small, round nose is located between the eyes. Mouth: A simple, curved line represents the mouth, which is smiling. Hair: The hair is depicted with several curved lines and loops, giving it a flowing and dynamic appearance. The hair extends outward from the face, creating a sense of movement and life. 3.~Additional Elements: Waves and Curves: The carving includes various waves and curves that flow around the face, adding to the dynamic and fluid nature of the design. These elements are integrated into the circular boundary, enhancing the overall aesthetic. Symmetry: The design is symmetrical, with the left and right sides of the face mirroring each other, contributing to a balanced and harmonious appearance. 4.~Texture and Surface: The carving is set against a rough, textured background, which appears to be a stone or concrete surface. 5.~Artistic Style: The style of the carving is reminiscent of ancient or traditional art forms, possibly inspired by symbols or faces found in historical or cultural contexts. The simplicity and elegance of the design suggest a focus on essential elements and a strong emphasis on symmetry and balance. In summary, the image shows a detailed stone carving of a stylized face or symbol, set against a textured background. The design is intricate, with elements such as eyes, nose, mouth, and flowing hair, all integrated into a circular boundary. The overall style is symmetrical and balanced, with a focus on simplicity and elegance.''}
        \end{subfigure}

        \medskip
        \begin{subfigure}{\linewidth}
            \caption{\textbf{CapMAS, 300 words.} ``The image depicts a circular stone carving or etching on a textured, light-colored surface. The carving features two intertwined figures, which appear to be stylized representations of human-like forms, with simplified facial features, including eyes, noses, and mouths. The figures are depicted in a flowing, interconnected manner, creating a sense of unity and harmony. The central part of the carving features two faces, each with eyes represented by small circles. The figures are adorned with intricate patterns and designs, including curved lines that form a pattern resembling a crown or a series of waves. These patterns add a decorative element to the carving. The background of the carving is a textured stone surface, which appears to be weathered and slightly rough, adding depth and contrast to the carving. The stone's color is a muted gray, which complements the darker lines of the carving, making the design more pronounced. The circular nature of the carving suggests a sense of unity and wholeness, as the two figures are seamlessly connected within the same space. The intertwined forms could symbolize various concepts such as unity, balance, or the interconnectedness of life. The simplicity of the design, combined with the rough texture of the stone, gives the carving a timeless and enduring quality. The stone surface on which the carving is etched has a rough texture, with visible grains and imperfections, adding a rustic and natural feel to the artwork. The contrast between the smooth lines of the carving and the rough texture of the stone surface creates a visually interesting and dynamic effect. The image presents a harmonious and artistic stone carving that combines abstract human figures with decorative patterns, set against a textured stone background. The design focuses on simplicity and elegance, making it a powerful and meaningful piece of art.''}
        \end{subfigure}

        \medskip
        \begin{subfigure}{\linewidth}
            \caption{\textbf{SimLoss FFT, 106 words.} ``The image shows an intricate circular design etched into a textured surface, likely concrete or stone. The design features two intertwined, stylized faces with exaggerated, flowing lines that give them a dynamic and almost animated appearance. Each face has a large, central eye and a smaller eye on the side, both depicted with simple, curved lines. The eyes are surrounded by elaborate, swirling patterns that extend outward, creating a sense of movement and fluidity. The overall effect is reminiscent of ancient or tribal art, with a focus on symmetry and organic forms. The texture of the surface adds depth to the design, making it appear three-dimensional.''}
        \end{subfigure}
    \end{minipage}
    \caption{\textbf{Caption length versus quality on \texttt{aar\_test\_04629}.} All three methods answer the same 18 of 24 questions. (a)~Plain VLM imposes a numbered breakdown and summary; (b)~CapMAS adds symbolic interpretation; (c)~SimLoss FFT uses roughly a third of the words and less interpretive padding, though it still speculates about artistic style.}
    \label{fig:qual_length_b}
\end{figure*}

Table~\ref{tab:results} shows that SimLoss FFT produces the shortest captions while achieving the highest precision. These two properties are not in tension; they reflect the same underlying property of the model. Figures~\ref{fig:qual_length_a} and~\ref{fig:qual_length_b} show two images on which all three models answer exactly the same set of questions -- so coverage is identical -- yet the captions differ dramatically in length and quality. In each figure the source image is shown on the left, with the caption produced by each method on the right.

On \texttt{aar\_test\_04779} (Figure~\ref{fig:qual_length_a}), all three models answer the same 20 out of 23 questions correctly. Plain VLM spends nearly a third of its words on a closing summary that repeats what was already said, and adds speculation about the setting (``possibly a campsite, a cabin'') with no visual basis. CapMAS infers the intended use of the wooden box (``likely used for storing or displaying items'') without evidence. SimLoss FFT covers the same questions in 123 words with less repetition and achieves 100\% measured precision against Plain VLM's 90.9\%, although its interpretation of the sign's tone illustrates that the judge can still accept subjective claims.

On \texttt{aar\_test\_04629} (Figure~\ref{fig:qual_length_b}), coverage is again identical across all models (18/24). Plain VLM structures its 395-word output as a numbered breakdown with headers and closes with a full summary paragraph, none of which adds answerable content. CapMAS at 300 words interprets the figures as symbols of ``unity, balance, or the interconnectedness of life,'' an inference that cannot be verified from the image. SimLoss FFT answers the same questions in 106 words and achieves 100\% precision against 94.1\% for Plain VLM and 89.7\% for CapMAS. Across these cases, the additional baseline words provide no measured coverage gain and coincide with lower precision; they are mainly summaries, speculation, and imposed structure.
\section{Prompts}
\label{sec:prompts}

This section lists the full set of prompts used in this work, verbatim. We group them into (i) the captioning prompts given to the VLM, (ii) the evaluation prompts used to compute precision, recall, and CLAIR on IIW-400, (iii) the inference-pipeline prompts of the CapMAS baseline, and (iv) the reward and judge prompts used by the reward-optimized baselines. Curly braces (e.g.\ \texttt{\{caption\}}) denote runtime substitutions, and \texttt{\textbackslash n} denotes a literal newline in a format-template prompt. All judge and evaluation calls use greedy decoding (temperature $\approx 0$).

\newenvironment{promptbox}{\par\smallskip\noindent\begin{quote}\small\ttfamily\raggedright}{\end{quote}\par\smallskip}

\subsection{Captioning Prompts}
\label{sec:prompts_caption}

\paragraph{Single-pass captioner.}
The single-pass captioning prompt across all methods is:
\begin{promptbox}
Describe this image in detail.
\end{promptbox}

\paragraph{CapMAS caption sampling.}
The CapMAS baseline first samples five captions per image from the VLM using five diverse detailed-description prompts:
\begin{promptbox}
1. Describe the given image in a very detailed manner.\\
2. Provide a detailed description of the specified image.\\
3. Elaborate on the details of the image provided.\\
4. Offer an in-depth description of the given image.\\
5. Thoroughly describe the features of the specified image.
\end{promptbox}

\subsection{Evaluation Protocol and Prompts}
\label{sec:prompts_eval}

Precision and recall are computed with a GPT-4o judge. Let $\mathcal{G}(\hat{y})$ be the atomic propositions extracted from candidate caption $\hat{y}$, and let $r$ be the IIW reference description. Define $J(g;v,r)=1$ when the judge finds proposition $g$ supported by image $v$ and description $r$, and 0 otherwise. Following atomic-fact factuality evaluation~\citep{min2023factscore,jing2024faithscore}, precision is
\begin{equation}
\mathrm{Precision}(\hat{y},v,r)
=
\frac{1}{|\mathcal{G}(\hat{y})|}
\sum_{g \in \mathcal{G}(\hat{y})}
J(g;v,r).
\end{equation}
For the human-verified question set $\mathcal{Q}(v)$, recall withholds the image and measures what can be recovered from the caption alone:
\begin{equation}
\mathrm{Recall}(\hat{y},v)
=
\frac{1}{|\mathcal{Q}(v)|}
\sum_{q \in \mathcal{Q}(v)}
\mathbf{1}[\textsc{Answer}(q\mid\hat{y}) \text{ is correct}].
\end{equation}
We report their harmonic mean, $\mathrm{F1}=2PR/(P+R)$. CLAIR compares the candidate caption with the IIW reference set using an LLM and requests a holistic same-image score from 0 to 100~\citep{chan-etal-2023-clair-evaluating}; Table~\ref{tab:results} reports this score divided by 100. The references and questions in this protocol are used only for evaluation, not as SimLoss adaptation targets. The exact prompts follow.

\paragraph{Proposition extraction (precision).}
The candidate caption is decomposed into atomic propositions (system prompt):
\begin{promptbox}
I want to verify if the given CAPTION is accurate. To assist with this verification, decompose the given CAPTION into atomic propositions. All parts of the caption must be broken down into propositions. The outputs should follow the following format: '1. proposition one\textbackslash n2. proposition two\textbackslash n3. proposition three'
\end{promptbox}

\paragraph{Proposition verification (precision).}
Each extracted proposition is judged True/False against the image and a reference caption; precision is the fraction judged True. System prompt:
\begin{promptbox}
Your role is to determine whether the given propositions are True or False based on the provided image and its description. The outputs should follow the following format: '1. True/False\textbackslash n2. True/False\textbackslash n3. True/False\textbackslash n4. True/False\textbackslash n5. True/False\textbackslash n ...'. The number of True/False answers must match the number of propositions.
\end{promptbox}
User message (with the image attached):
\begin{promptbox}
Description: \{reference\_caption\}\\[2pt]
Propositions:\\
\{propositions\}
\end{promptbox}

\paragraph{Question answering (recall).}
Recall is measured by answering the IIW-400 multiple-choice questions for each image using only the candidate caption, then comparing against the image-derived answer key. System prompt:
\begin{promptbox}
Your role is to answer the given questions based on the provided caption. I want to measure the amount of information in the caption. Therefore, if the correct answer to the question cannot be determined from the caption, you should answer it with "I don't know". Do not use your own knowledge in your response. Do not use information that can be inferred from the question itself. Only use the information provided in the caption. Answer the question by directly selecting the letter of the corresponding option. Do not repeat the question.
\end{promptbox}
User message:
\begin{promptbox}
Caption: \{caption\}.\\
Questions:\\
\{questions\}
\end{promptbox}

\paragraph{CLAIR.}
The CLAIR score is obtained by prompting the judge to rate, on a 0--100 scale, how likely the candidate and reference caption sets describe the same image:
\begin{promptbox}
You are trying to tell if a candidate set of captions is describing the same image as a reference set of captions.\\
Candidate set:\\
\{candidate\_statements\}\\
Reference set:\\
\{target\_statements\}\\
On a precise scale from 0 to 100, how likely is it that the candidate set is describing the same image as the reference set? (JSON format, with a key "score", value between 0 and 100, and a key "reason" with a string value.)
\end{promptbox}

\subsection{CapMAS Inference-Pipeline Prompts}
\label{sec:prompts_capmas}

After sampling the five captions above, CapMAS merges, decomposes, fact-checks, and rewrites them.

\paragraph{Stage 2 --- Merge (LLM).}
\begin{promptbox}
You are an expert at merging multiple image captions into a single, comprehensive description. Based on the following captions describing the same image, create ONE detailed merged caption that incorporates all the important details mentioned across the different captions. Be thorough but avoid repetition. Only describe what is actually mentioned in the captions.
\end{promptbox}

\paragraph{Stage 3 --- Decompose into propositions (LLM).}
\begin{promptbox}
I want to verify if the given CAPTION is accurate. To assist with this verification, decompose the given CAPTION into atomic propositions. All parts of the caption must be broken down into propositions. The outputs should follow the following format: '1. proposition one\textbackslash n2. proposition two\textbackslash n3. proposition three'. For example, break down 'He is tall, thin, and pale' into '1. He is tall.\textbackslash n2. He is thin.\textbackslash n3. He is pale.'
\end{promptbox}

\paragraph{Stage 4 --- Visual fact-check (VLM).}
Each proposition is fact-checked by querying the VLM with the image and the prompt below, and reading the first-token log-probabilities of \texttt{True} and \texttt{False}. A proposition is kept when its hallucination score $h = -\log(\max(p_{\mathrm{true}}-p_{\mathrm{false}}, \epsilon))$ falls below a threshold.
\begin{promptbox}
True or False? \{proposition\}
\end{promptbox}

\paragraph{Stage 5 --- Rewrite from verified facts (LLM).}
\begin{promptbox}
I want to create a caption that includes only facts. Please help me correct the given caption. The given caption contains things that are not true. Based on the given FACTS and NON-FACTS, remove the non-factual elements from the caption. Place the revised caption between '\#\#\#'.
\end{promptbox}

\subsection{Reward and Judge Prompts}
\label{sec:prompts_reward}

The FeedQuill PPO baseline first breaks a generated caption into smaller, atomic information units. A judge scores these units to obtain unit-level precision and recall, and their harmonic mean is used as the F1 component $R_{\mathrm{DC}}$ of the reward in Section~\ref{sec:reward}. PAPO instead uses a perception-consistency loss (the KL divergence between the caption distributions on the original and masked images) and therefore introduces no additional judge prompt. SimLoss FFT uses a frozen-encoder embedding-alignment objective and SimLoss GRPO uses a black-box image--text embedding-similarity reward (Gemini~2~Embed); neither uses a text judge prompt.

\paragraph{Caption decomposition.}
The generated caption is decomposed into short, single-fact propositions:
\begin{promptbox}
Decompose the following image caption into atomic propositions.\\[2pt]
Rules:\\
- Each proposition must be a single, verifiable visual fact (one subject + one predicate).\\
- Keep propositions SHORT (<= 12 words each).\\
- Do NOT split facts that are inherently linked (e.g.\ "red car on the left" -> one prop).\\
- Do NOT add information not in the caption.\\
- Do NOT include vague or subjective claims (e.g.\ "the scene looks peaceful").\\
- Include spatial relations, colors, counts, object states, and text if present.\\
- Aim for 5-15 propositions depending on caption length.\\[2pt]
Caption:\\
"\{caption\}"\\[2pt]
Respond ONLY with a JSON object, no markdown, no extra text:\\
\{\{"propositions": ["prop1", "prop2", "prop3", ...]\}\}
\end{promptbox}

\paragraph{Unit verification.}
Each proposition is verified against the image:
\begin{promptbox}
Look at the image carefully.\\[2pt]
Proposition: "\{proposition\}"\\[2pt]
Is this proposition TRUE or FALSE based solely on what you can see in the image?\\
- Reply TRUE if the proposition is clearly visible and correct.\\
- Reply FALSE if it is wrong, hallucinated, or not visible.\\
- Do NOT use external knowledge. Judge only from the image.\\[2pt]
Reply with ONLY the word TRUE or FALSE.
\end{promptbox}
For efficiency, propositions from all candidate captions for an image are verified in a single vision call using a batched variant that asks for a JSON map from each proposition key to \texttt{true}/\texttt{false}.

\paragraph{Unit-level F1 reward.}
The judge decisions over the decomposed units are aggregated into unit-level precision $P$ and recall $R$. The FeedQuill score is their harmonic mean, $R_{\mathrm{DC}}=2PR/(P+R)$.

\end{document}